\documentclass[journal,twoside,web]{ieeecolor}
\usepackage{generic}
\usepackage{cite}
\usepackage{amsmath,amssymb,amsfonts,bm}
\usepackage{algorithm}
\usepackage{algorithmic}
\usepackage{graphicx}
\usepackage{epstopdf-base} % Use the supplied logo PDF even when shell escape is disabled.
\usepackage{textcomp}

\usepackage{array}
\usepackage{booktabs}
\usepackage{colortbl}
\usepackage{tabularx}
\usepackage{multirow}
\usepackage{makecell}
\usepackage{threeparttable}
\usepackage{siunitx}
\usepackage{float}
\usepackage{placeins}
\usepackage{gensymb}
\usepackage{pifont}
\usepackage{url}

\definecolor{oursrow}{RGB}{242,245,249}
\definecolor{bestrow}{RGB}{236,245,242}
\definecolor{sectbg}{RGB}{246,247,249}
\definecolor{gainc}{RGB}{21,128,61}
\newcommand{\gain}[1]{\textcolor{gainc}{\scriptsize(#1)}}
\newcolumntype{C}{>{\centering\arraybackslash}X}

\usepackage{hyperref}
\hypersetup{
    colorlinks=false,
    pdfborder={0 0 0},
    pdfstartview=Fit,
    breaklinks=true,
    pdftitle={EchoMERIT: Unifying Clinical Experience, Tool Correction, and Deterministic Verification for Echocardiography Agents},
    pdfauthor={Xinyue Liang, Yue Zhou, Dianye Huang, Zhongliang Jiang}
}

\graphicspath{{figures/}{figures/predictor/}}

\newcommand{\sysname}{\textsc{EchoMERIT}}
\newcommand{\bench}{\textsc{EchoToolBench}}

\def\BibTeX{{\rm B\kern-.05em{\sc i\kern-.025em b}\kern-.08em
    T\kern-.1667em\lower.7ex\hbox{E}\kern-.125emX}}
\begin{document}
\bstctlcite{IEEEexample:BSTcontrol}

\title{EchoMERIT: Unifying Clinical Experience, Tool Correction, and Deterministic Verification for Echocardiography Agents}

\author{\mbox{Xinyue Liang, Yue Zhou, Dianye Huang, and Zhongliang Jiang}%
\thanks{Xinyue Liang, Dianye Huang, and Zhongliang Jiang are with The University of Hong Kong, Hong Kong. Yue Zhou is with the Technical University of Munich, Germany.}%
\thanks{Corresponding author: Zhongliang Jiang (e-mail: \mbox{zljiang@hku.hk}).}%
\thanks{Project: \url{https://github.com/xy-liang22/EchoMERIT}.}%
\endgraf
}

\maketitle

\begin{abstract}
Quantitative echocardiography requires synthesizing multiple cine views, executing guideline equations, and deferring when an exam cannot support a query. Multimodal large language models answer coarse visual questions, but end-to-end measurement regression and formula chaining remain unreliable. Tool-augmented agents delegate measurement to specialist cardiac encoders, yet frozen encoders drift and conflict unnoticed, clinical experience is rarely reusable across planners, and stochastic LLM critics miss deterministic execution errors. We present EchoMERIT (Memory-Enhanced Reasoning with Integrated Tool Verification), a closed-loop echocardiography agent framework with three components. An asymmetric deterministic verification harness audits tool outputs, physiological consistency, formulas, units, and evidence citations through bounded self-correction, refuses out-of-scope queries, and supplies process signals for policy learning, all without an auxiliary verifier LLM. Embedded in the harness, attribution-conditioned visual error calibration (Attr-Corr) pairs frozen-encoder gradient attribution maps with raw predictions to correct residual errors without retraining foundation models. A read-only clinical experience memory distills calibration priors and workflow playbooks offline. On a four-scope benchmark we construct from MIMIC-IV-Echo and EchoNet, these modules improve local and frontier planners without weight updates; with verification-guided training, an 8B planner rises from 56.6\% to 70.3\% concept-grounded Study accuracy, matching frontier configurations, while invalid tool calls drop from 0.56 to 0.01 per query. On EchoNet-Dynamic, Attr-Corr outperforms scalar and shuffled-attribution controls and reduces ejection-fraction error $5.2\times$ more than LoRA-tuned VLMs. EchoMERIT further generalizes to held-out tasks while reliably refusing unsupported queries.
\end{abstract}


% Main body (strictly <= 8.0 pages)
% ==============================================================================
% Section I: Introduction (V18 paradigm-level narrative; work-specific details in Sec. II)
% ==============================================================================
\section{Introduction}
\label{sec:intro}

\begin{figure*}[t]
\centering
\includegraphics[width=\textwidth]{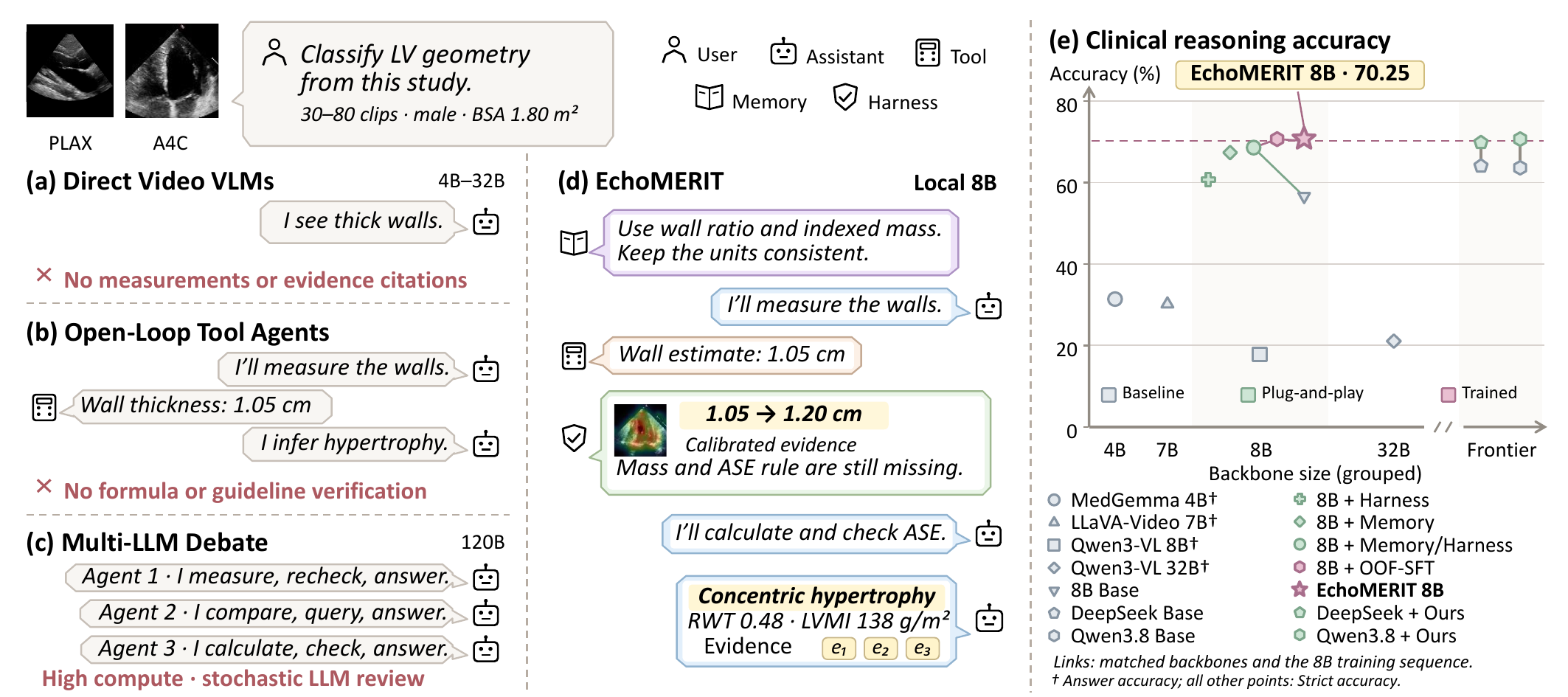}
\caption{Paradigm comparison on multi-step quantitative echocardiography. (a)--(d)~Illustrative left-ventricular (LV) geometry workflow contrasting direct video VLMs, open-loop tool agents, $120\text{B}$ multi-LLM debate, and \sysname{} on a local $8\text{B}$ backbone (clinical memory, attribution-conditioned visual calibration, and deterministic verification emitting auditable evidence handles $[e_X]$); ultrasound frames are de-identified illustrations. (e)~Multi-view Study blind-test accuracy grouped by baseline, plug-and-play, and trained configurations ($N{=}242$). $^{\dagger}$Direct VLMs are plotted with Answer accuracy because they cannot cite evidence; all tool-augmented agents use Concept-Grounded Strict accuracy, so the two groups are not directly comparable.}
\label{fig:paradigm}
\end{figure*}

\IEEEPARstart{T}{ransthoracic} echocardiography is the primary bedside modality for assessing cardiac structure and hemodynamics. A routine quantitative report requires synthesizing multiple cine views, measuring chamber dimensions and Doppler velocities, applying sex-specific guideline equations~\cite{lang2015chamber,baumgartner2017aortic}, and deferring whenever an exam cannot support a query. Cardiac AI has progressed from task-specific models that automate single measurements on individual clips~\cite{ouyang2020echonetdynamic,duffy2022echonetlvh,sahashi2025echonetmeasurements} to study-level foundation models that embed or predict dozens of phenotypes across views~\cite{christensen2024echoclip,holste2025panecho,vukadinovic2025echoprime}. These models are accurate within their trained outputs, yet they cannot chain downstream guideline formulas, answer queries beyond their fixed heads, ground answers in auditable evidence, or recognize their own estimation errors.

Multimodal large language models (MLLMs) promise to replace fixed heads with flexible clinical reasoning. Direct general and medical video VLMs~\cite{moor2023medflamingo,li2023llavamed,zhang2024biomedgpt,wu2025radfm,sellergren2026medgemma,sellergren2026medgemma15,zhang2024llavavideo,bai2025qwen3vl} answer queries from sampled frames (Fig.~\ref{fig:paradigm}(a)), but asking one end-to-end model to regress continuous measurements and chain physiological equations is unreliable and unverifiable. Tool agents instead let an LLM planner call specialist encoders and calculators~\cite{suris2023vipergpt,wang2024videoagent,fallahpour2025medrax,lou2026cxragent,daghyani2026echoagent} (Fig.~\ref{fig:paradigm}(b)), but they remain open-loop in the sense that no dedicated layer verifies tool outputs or execution. Tool access removes direct regression, but it does not tell the planner how the tools behave in practice, so planners still misread raw probabilities, skip intermediate formulas, and confuse units. Current remedies store static declarative facts, fine-tune experience into planner weights at the risk of overfitting to seen tasks, or maintain mutable memories that depend on post-episode feedback~\cite{zhao2024expel,xiao2025toolmem,li2026hepm,shen2026evomedagent}. What is still missing is read-only, scope-aware clinical experience that any planner can reuse.

To improve reliability, recent frameworks add multi-perspective LLM discussion or auxiliary LLM reviewers on top of the planner~\cite{heidari2025echocopilot,zhang2026radagents} (Fig.~\ref{fig:paradigm}(c)). Such critics share the stochastic failure modes of the planner they review, so they can overlook a unit swap, an omitted equation, or a mismatched citation while multiplying compute. Quantitative echocardiography, however, follows standardized procedures of measurement, formula calculation, guideline lookup, and scope checking, which means most execution errors can be caught by a program rather than by another language model. A deeper bottleneck lies upstream. Every tool-augmented agent is bounded by its frozen specialist tools, which exhibit systematic numerical drift, borderline threshold instability, and cross-tool contradictions~\cite{heidari2025echocopilot,fallahpour2025medrax,huai2026teacxa,gan2026csrl}. Seeing only raw scalar outputs, a planner cannot tell when an encoder has erred, while retraining foundation models for each site is costly and a secondary VLM critic is slow and noisy.

We propose \sysname{} (\emph{Memory-Enhanced Reasoning with Integrated Tool Verification}), a closed-loop agent framework for auditable quantitative echocardiography (Fig.~\ref{fig:paradigm}(d) and Fig.~\ref{fig:overview}). Its core is an asymmetric deterministic verification harness that replaces stochastic LLM critics with fast, non-LLM checks: recoverable omissions trigger bounded diagnostic retries, whereas out-of-scope queries trigger hard refusal. Inside the harness, an attribution-conditioned visual error calibrator reads the gradient attribution maps of frozen encoders alongside their raw predictions, so that drift is detected and corrected before it propagates into downstream formulas. A read-only clinical experience memory supplies distilled calibration priors and workflow playbooks to whichever planner is deployed. Because all three modules operate outside planner weights, they transfer plug-and-play across local and frontier backbones, and the harness signals can further guide offline policy learning on a compact $8\text{B}$ planner~\cite{yang2025qwen3}. Our contributions are threefold:
\begin{itemize}
\item \textbf{Asymmetric Deterministic Verification Harness.} A lightweight non-LLM verifier that audits tool outputs, physiological consistency, formulas, units, and evidence citations through bounded self-correction, enforces hard refusal on out-of-scope queries, and provides deterministic process signals for local policy learning.
\item \textbf{Attribution-Conditioned Visual Error Calibration.} \emph{Attr-Corr}, an error sensor embedded in the harness that jointly reads gradient attribution maps and raw predictions of frozen cardiac encoders to estimate residual corrections and risk, mitigating upstream drift and conflicts without retraining foundation models.
\item \textbf{Read-Only Clinical Experience Memory.} An offline-distilled, scope-partitioned memory of calibration priors and workflow playbooks that transfers plug-and-play across local and frontier planners without updating their weights.
\end{itemize}

% ==============================================================================
% Section II: Related Work (V18: work-specific comparisons moved here from Sec. I)
% ==============================================================================
\section{Related Work}
\label{sec:related_work}

\subsection{Echocardiography Models and Direct Multimodal Answering}
\label{subsec:rw_fm_agents}
Task-specific video models such as EchoNet-Dynamic~\cite{ouyang2020echonetdynamic}, EchoNet-LVH~\cite{duffy2022echonetlvh}, and EchoNet-Measurements~\cite{sahashi2025echonetmeasurements} automate single-clip ejection fraction and caliper quantification. Study-level foundation models broaden coverage in two ways: EchoCLIP~\cite{christensen2024echoclip} learns contrastive video--report embeddings for retrieval, whereas PanEcho~\cite{holste2025panecho} and EchoPrime~\cite{vukadinovic2025echoprime} predict dozens of phenotypes across views through fixed output heads. Because these encoders are accurate within their trained outputs, \sysname{} keeps them frozen and treats them as tools. General and biomedical VLMs, including Med-Flamingo~\cite{moor2023medflamingo}, LLaVA-Med~\cite{li2023llavamed}, BiomedGPT~\cite{zhang2024biomedgpt}, RadFM~\cite{wu2025radfm}, MedGemma~\cite{sellergren2026medgemma,sellergren2026medgemma15}, LLaVA-Video~\cite{zhang2024llavavideo}, and Qwen3-VL~\cite{bai2025qwen3vl}, extend open-ended multimodal question answering, and grounded radiology pipelines such as MAIRA-2~\cite{bannur2024maira2} and LoBA~\cite{nguyen2025loba} localize findings before answering. Without tools, however, these models struggle with multi-view quantitative calculation, guideline synthesis, and explicit coverage boundaries.

\subsection{Clinical Tool Agents and Multi-Agent Verification}
\label{subsec:rw_agents}
Tool-augmented agents decompose visual reasoning into calls to specialist models~\cite{suris2023vipergpt,wu2024vstar,wang2024videoagent}. In clinical imaging, MedRAX~\cite{fallahpour2025medrax} orchestrates chest X-ray tools in a ReAct loop with short-term execution memory, CXRAgent~\cite{lou2026cxragent} composes director-orchestrated multi-stage workflows, and MedScope~\cite{li2026medscope} learns coarse-to-fine tool calling over clinical videos. In echocardiography, EchoAgent~\cite{daghyani2026echoagent} couples open planners from $8\text{B}$ to $30\text{B}$ with iterative feedback, feasibility filters, and guideline retrieval. Without a dedicated verification layer, such agents largely trust raw tool outputs, although tool reliability is a known weak point: a tool-expertise-aware chest X-ray agent~\cite{huai2026teacxa} learns which conflicting tool response to trust, and Gan et al.~\cite{gan2026csrl} show that tool failures can prevent medical agents from realizing synergistic tool gains. A complementary line adds multi-agent review. RadAgents~\cite{zhang2026radagents} resolves conflicts in radiology reasoning through multi-agent grounding and retrieval, and Echo-CoPilot~\cite{heidari2025echocopilot} runs a three-perspective ReAct discussion on a $120\text{B}$ planner, supported by an echocardiography knowledge graph (EchoKG) and LLM self-contrast. These reviewers remain stochastic and compute-intensive, and a knowledge graph records static facts rather than how a particular tool chain fails. \sysname{} instead runs on a local $8\text{B}$ backbone, checks citations, formulas, and guideline rules deterministically against society standards~\cite{lang2015chamber,baumgartner2017aortic,nagueh2016diastolic,heidenreich2022hf,mukherjee2025rightheart} without an auxiliary LLM, calibrates tool outputs from attribution maps, and reuses verification signals for policy learning.

\subsection{Clinical Memory, Guardrails, and Visual Error Sensing}
\label{subsec:rw_reliability_memory}
Selective prediction~\cite{geifman2019selectivenet}, conformal risk control~\cite{angelopoulos2024conformal}, reliability stress testing~\cite{yang2026reliability}, and cardiac guardrails that predict LVEF error risk for deferral~\cite{kabra2026guardrails} mitigate unreliable outputs, yet deferral cannot correct drift when a single frozen encoder covers a measurement. In agentic memory, ExpeL~\cite{zhao2024expel} and ToolMem~\cite{xiao2025toolmem} extract insights or tool capabilities for test-time retrieval, H-EPM~\cite{li2026hepm} pairs hybrid episodic-procedural memory with RL, and Evo-MedAgent~\cite{shen2026evomedagent} updates inter-case memory from post-episode feedback. For frozen clinical deployment without per-case labels, \sysname{} instead combines scope-partitioned calibration priors with read-only playbooks. Gradient attribution maps~\cite{selvaraju2017gradcam} are usually audited post hoc for faithfulness or shortcut reliance~\cite{adebayo2018sanity,geirhos2020shortcut,zafar2026visualreliance}; \sysname{} repurposes them as fast error sensors and validates them against shuffled-map controls.

\subsection{Verifiable Multi-Turn Agentic Reinforcement Learning}
\label{subsec:rw_rl}
Offline preference optimization~\cite{rafailov2023dpo} and verifiable RL via GRPO~\cite{shao2024deepseekmath} with Schulman $k_3$ KL regularization~\cite{schulman2020klapprox}, refined by Dr.\ GRPO~\cite{liu2025drgrpo} and DAPO~\cite{yu2025dapo}, have advanced reasoning models and recently extended to video agents~\cite{zhang2026eva}, visual tool diversity~\cite{kim2026diversity}, adaptive stopping~\cite{zhou2026adaptivestopping}, and failure-aware medical tool use~\cite{gan2026csrl}. Shaped rewards, however, can destabilize GRPO on all-fail groups through variance normalization~\cite{wang2026darkroom} or create phantom advantages under shaped-score filtering~\cite{yu2026filtermetric}. For multi-turn clinical tool use, where sparse $0/1$ rewards erode tool syntax and ungated bonuses reward wrong rollouts, \sysname{} pairs offline training labels with deterministic harness signals in an outcome-gated reward and clamps the advantages of wrong rollouts.

% ==============================================================================
% Section III: Method (V18 academic rewrite: no code identifiers, Fig. 2-aligned notation)
% ==============================================================================
\section{Method}
\label{sec:method}

\begin{figure*}[t]
\centering
\includegraphics[width=\textwidth]{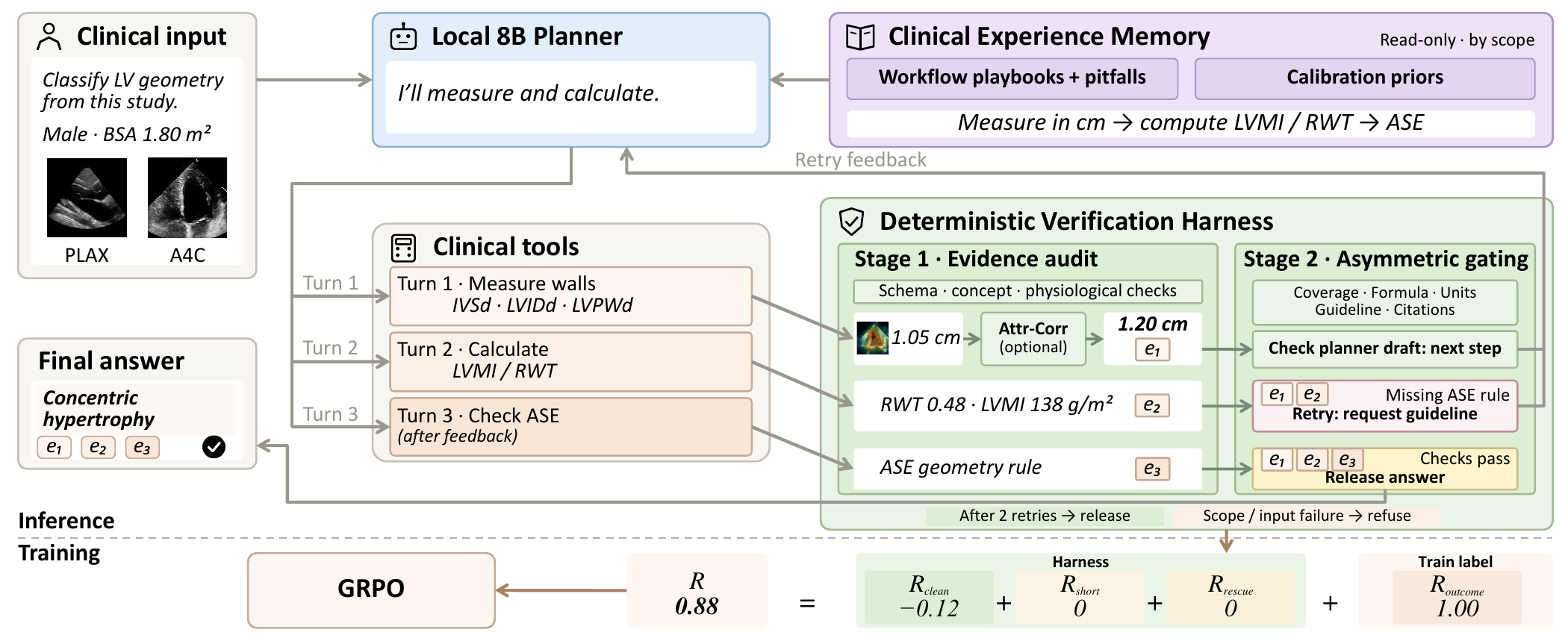}
\caption{Architecture of \sysname{} illustrated on an LV geometry study. \emph{Top (Inference):} read-only scope-partitioned clinical memory guides a local $8\text{B}$ planner (frontier planners use the same loop) across multi-turn clinical tools, while a deterministic verification harness (without an auxiliary LLM) audits each turn's evidence record ($[e_1]$--$[e_3]$, with embedded \emph{Attr-Corr} visual calibration) and enforces bounded retries or hard out-of-scope refusal. \emph{Bottom (Offline Training):} deterministic harness process terms ($R_{\text{clean}}, R_{\text{short}}, R_{\text{rescue}}$) pair with offline training reference labels ($R_{\text{outcome}}$) to drive multi-turn GRPO following supervised warm-start (illustrated for a $1$-retry strictly correct rollout, $R{=}0.88$). Ultrasound frames are de-identified illustrations.}
\label{fig:overview}
\end{figure*}

\subsection{Problem Formulation and Overview}
\label{subsec:method_overview}
\sysname{} casts quantitative echocardiography as a verified multi-turn decision process. An LLM planner gathers evidence from clinical tools, while deterministic modules outside the planner regulate which evidence it sees and which answers it may release (Fig.~\ref{fig:overview}).

Each episode receives a clinical query $q$, optional answer choices $\mathcal{O}$, and an input $x$ from one of four scopes $s\in\mathcal{S}=\{\text{Study},\text{Single-Video},\text{Spatial},\text{Text}\}$. A Study input is a full multi-view examination, a Single-Video input is one cine clip, a Spatial query asks for pixel-level outputs on a clip, and a Text input provides measurements without images. The planner $\pi_\theta$ acts on a typed tool set $\mathcal{T}$ organized into five semantic families. Multi-view cardiac encoders expose the $40$ heads of PanEcho and the $21$ outputs of EchoPrime~\cite{holste2025panecho,vukadinovic2025echoprime}. Single-clip spatial specialists built on EchoNet-Dynamic and EchoNet-LVH~\cite{ouyang2020echonetdynamic,duffy2022echonetlvh} return ventricular masks, caliper keypoints, and ejection fraction. A physiological formula calculator implements $16$ deterministic equations, and a clinical guideline rule engine encodes $19$ rules derived from society standards~\cite{lang2015chamber,baumgartner2017aortic,nagueh2016diastolic,zoghbi2017regurgitation,heidenreich2022hf,mukherjee2025rightheart,nagueh2025diastolicupdate} and documented reporting cutoffs. Finally, a capability registry reports the available views and the supported query scope. Concrete tool interfaces are listed in Appendix~\ref{app:system_specs}.

A trajectory $\tau=(a_1,o_1,\dots,a_T,o_T,\hat{y})$ records the interaction, where each action $a_t$ is a tool call or a final response and each observation $o_t$ is the returned tool result. The planner runs at most eight main-loop turns, optionally followed by one synthesis turn, and dispatches at most twelve tool calls. Every executed call appends a structured record, holding the clinical concept, value, unit, and source tool, to an evidence store $\mathcal{E}=\{e_1,\dots,e_k\}$. The episode ends either with an answer $\hat{y}$ that cites evidence handles $[e_X]$ or with an explicit abstention. We score it with concept-grounded strict accuracy
\begin{equation*}
y_{\text{strict}}=\mathbb{I}\big[\hat{y}\simeq y^{\star}\big]\cdot\mathbb{I}\big[\exists\,e\in\mathcal{C}(\hat{y}):\ \kappa(e)\in\mathcal{K}_q\big],
\end{equation*}
where $\hat{y}\simeq y^{\star}$ denotes a match with the reference within the task tolerance, $\mathcal{C}(\hat{y})\subseteq\mathcal{E}$ is the set of cited records, $\kappa(e)$ is the clinical concept of record $e$, and $\mathcal{K}_q$ is the set of concepts relevant to the query. A correct answer therefore counts only when it rests on the right evidence. For unsupported queries, abstention is the only correct response.

Around this loop, the read-only clinical memory (Section~\ref{subsec:method_memory}) conditions the planner before it acts, the deterministic harness (Section~\ref{subsec:method_harness}) audits every observation and draft answer with \emph{Attr-Corr} (Section~\ref{subsec:method_error_predictor}) embedded inside it, and optional policy learning (Section~\ref{subsec:method_posttrain}) internalizes the harness signals into a local planner.

\subsection{Scope-Partitioned Clinical Experience Memory}
\label{subsec:method_memory}
General-purpose planners that call medical tools frequently misinterpret raw encoder probabilities, skip intermediate formulas, or confuse centimeters with millimeters. A knowledge graph of textbook facts~\cite{heidari2025echocopilot} does not tell a planner how its own tools fail, fine-tuning such lessons into weights risks overfitting to seen tasks, and test-time mutable memories~\cite{shen2026evomedagent} require post-episode labels and let behavior drift across patients.

Building on offline experience extraction~\cite{zhao2024expel,xiao2025toolmem}, \sysname{} distills execution behavior on training patients into a compact memory that is frozen at deployment. The memory contains no patient identifiers and is partitioned by input scope, so Study and Single-Video queries retrieve guidance fitted to their own type of evidence. It consists of two layers. The first layer holds calibration priors. For each encoder head, we fit a decision threshold on training patients and mark a borderline band around it, inside which the planner is advised to seek corroborating evidence. Heads with weak training discrimination are flagged so that the planner does not over-trust their raw probabilities. The second layer holds workflow playbooks and error lessons, which are short multi-step recipes and pitfall warnings distilled from training error analyses. For example, the LV geometry playbook instructs the planner to measure wall thickness in centimeters, compute relative wall thickness and indexed mass, and then apply the ASE geometry rule. At inference, lexical and pattern matching over the query selects the relevant entries and prepends a compact memory card to the planner context (Appendix~\ref{app:system_specs}). Because the memory is read-only and external to the planner, it introduces no test-time drift, needs no per-case feedback, and transfers unchanged across local and frontier backbones.

\subsection{Asymmetric Deterministic Verification Harness}
\label{subsec:method_harness}
Prompting and memory cannot guarantee disciplined execution, and an auxiliary verifier LLM adds stochastic blind spots and latency. Quantitative echocardiography, however, follows standardized procedural rules that a program can check. \sysname{} therefore routes every tool response and every draft answer through a two-stage deterministic harness that makes no LLM calls and whose rule checks run in under a millisecond. The complete rule and gate registries are given in Appendix Table~\ref{tab:app_rules_gates}.

\textbf{Stage~1: Evidence audit.} As each tool returns, the harness validates the call arguments and the output schema. It then invokes \emph{Attr-Corr} on supported encoder heads (Section~\ref{subsec:method_error_predictor}) and checks five physiological consistency pairs, such as an ejection-fraction disagreement of at least $10$ percentage points between two encoders. It also flags anatomical concept substitutions, for instance citing tricuspid-valve evidence for a pulmonic-valve query. Each finding is attached to the corresponding evidence record.

\textbf{Stage~2: Asymmetric gating.} Before a draft answer is released, eight deterministic gate families sort execution errors by clinical risk and respond asymmetrically. Recoverable errors receive bounded self-correction. These include omitted formulas or guideline checks, missing or invalid evidence handles, malformed multiple-choice answers, centimeter--millimeter swaps, and premature abstentions. For such errors, Gates~2--8b return targeted diagnostic feedback, such as a reminder that the ASE guideline check is missing, and allow up to two retries. If the retry budget is exhausted, the draft is released and the event is logged, so recovery is encouraged but not guaranteed. Scope violations instead receive hard refusal. When a query targets an uncovered pathology or lacks the required inputs, Gates~1 and~1b override the draft with a forced abstention. This asymmetry keeps the agent helpful on recoverable mistakes while refusing deterministically whenever the scope check fails. Every gate event and tool error is also logged, and these free deterministic signals later serve as reward terms for policy learning (Section~\ref{subsec:method_posttrain}).

\subsection{Attribution-Conditioned Visual Error Calibration}
\label{subsec:method_error_predictor}
Even flawless orchestration is capped by the frozen encoders, whose outputs drift systematically and can disagree across tools~\cite{heidari2025echocopilot,huai2026teacxa,gan2026csrl}. Retraining foundation encoders for each site is expensive, and a fine-tuned VLM critic is slow and can even increase error. Our key observation is that each encoder pass already yields spatial evidence associated with its prediction: the gradient attribution map highlights the regions that drove the output, and misplaced attribution can accompany an unreliable estimate.

\emph{Attr-Corr} exploits this signal as a lightweight error sensor $f_\phi$ embedded in Stage~1 of the harness. For an encoder head $h$, it reads the raw prediction $\hat{y}_h$ together with the gradient attribution maps $H$~\cite{selvaraju2017gradcam} extracted during the same encoder pass, and it predicts a residual correction and a failure-risk score, $f_\phi(\hat{y}_h,H)\to(\Delta\hat{y}_h,r_h)$. We instantiate two backends matched to the available data, with training objectives given in Appendix~\ref{app:system_specs}. The \emph{3D-Attr} backend is an R(2+1)D-18 network~\cite{tran2018r21d} with BiomedBERT~\cite{gu2021biomedbert} FiLM conditioning on the task description. It consumes ten spatiotemporal channels that combine visual, segmentation, and attribution inputs, and it serves the dynamic and spatial ejection-fraction tasks. The \emph{ZeroRGB-Attr} backend is a five-fold ResNet-18 ensemble~\cite{he2016resnet} whose five input channels contain attribution maps with the RGB pixels zeroed out. Removing raw pixels limits shortcut learning on the small MIMIC training cohort, and this backend covers $60$ clinical heads for the online Study and Single-Video agents.

At inference, fold $k$ of the ensemble proposes a corrected value $c_{h,k}=\hat{y}_h+\Delta\hat{y}_h^{(k)}$, and $r$ denotes the mean failure risk across folds. The calibrated output and the risk alert are
\begin{equation}
\label{eq:error_pred_gate}
\begin{aligned}
\tilde{y}_h &=
\begin{cases}
\hat{y}_h, & \tau_h > 0 \land r < \tau_h, \\
\operatorname{agg}_{k=1..5}\, c_{h,k}, & \text{otherwise},
\end{cases} \\
\hat{y}_{\text{corr}} &= \operatorname{clip}(\tilde{y}_h,\ell_h,u_h), \\
h_{\text{risk}} &= \mathbb{I}\!\big[r \ge 0.75 \;\lor\; \text{agree} < 0.79\big],
\end{aligned}
\end{equation}
where $[\ell_h,u_h]$ are physiological bounds and $\text{agree}$ is the fraction of folds that agree with the ensemble decision. Regression heads use $\tau_h{=}0$ and aggregate by the fold median, so they are always calibrated. Binary and three-class heads aggregate by the fold mean of class probabilities and keep the raw output whenever the risk falls below an out-of-fold threshold $\tau_h{=}\tau_{\text{type}}$ calibrated per task type. Heads without a trained sensor and heads flagged as unreliable by the memory are left untouched.

Accepted corrections are written back in two tracks. The raw encoder prediction is archived in $\mathcal{E}$ for clinical audit, while the calibrated value and any alert $h_{\text{risk}}$ are exposed to downstream formulas, guideline rules, and the planner. Calibration thus intervenes before errors propagate through formula chains, adds about $18$\,ms per query, and leaves the foundation encoders unchanged.

\subsection{Verification-Guided Policy Learning}
\label{subsec:method_posttrain}
Memory and verification already help any planner without weight updates, but a compact $8\text{B}$ planner still wastes turns on redundant or malformed calls. We therefore optionally internalize disciplined execution into the local planner without querying an external teacher. Multi-turn RL must then avoid two known pitfalls: sparse $0/1$ outcome rewards erode tool syntax, whereas ungated shaped rewards can push the policy toward incorrect trajectories.

\textbf{Supervised warm-start.} We fine-tune a LoRA adapter~\cite{hu2022lora} for one epoch on $1{,}475$ imaging trajectories that combine programmatic workflows with strictly correct, error-free rollouts of the $8\text{B}$ planner. The trajectories are constructed out of fold across training patients, so no trajectory sees memory or calibration fitted on its own patient. Text-scope tasks are excluded entirely, which lets the Text evaluation measure cross-scope transfer directly.

\textbf{Verification-guided GRPO.} Starting from the warm-start policy, we optimize multi-turn tool use with GRPO~\cite{shao2024deepseekmath,liu2025drgrpo,yu2025dapo}. For each rollout $\tau_i$ in a group of $G$ rollouts, offline training labels determine strict correctness $y_{\text{strict},i}$, and the harness supplies the process terms. The reward follows the decomposition in Fig.~\ref{fig:overview}:
\begin{equation}
\label{eq:reward_fn}
R(\tau_i) =
\begin{cases}
\begin{aligned}
&\max\!\big(0.80,\; R_{\text{outcome},i} + R_{\text{clean},i} \\[-2pt]
&\quad + R_{\text{short},i} + R_{\text{rescue},i}\big),
\end{aligned} & y_{\text{strict},i}=1, \\[8pt]
\begin{aligned}
&\operatorname{clip}\!\big(R_{\text{soft},i} - P_{\text{clean},i} \\[-2pt]
&\quad - P_{\text{safety},i},\; -0.60,\; 0.15\big),
\end{aligned} & y_{\text{strict},i}=0.
\end{cases}
\end{equation}
For a strictly correct rollout, the outcome term $R_{\text{outcome},i}{=}1$ is adjusted by three harness terms. The cleanliness term $R_{\text{clean},i}$ grants a bonus for execution without retries or tool errors and otherwise subtracts a penalty for each gate retry and tool error. The brevity term $R_{\text{short},i}$ favors the shortest call chain among spotless correct peers in the same group. The rescue term $R_{\text{rescue},i}$ grants a small credit when a clean cross-checking rollout succeeds in a group whose non-cross-checking peers all fail. The rollout in Fig.~\ref{fig:overview}, for example, needs one retry and receives $R=1.00-0.12=0.88$. For an incorrect rollout, a bounded partial-credit term $R_{\text{soft},i}$ is reduced by an execution penalty $P_{\text{clean},i}$ for gate retries, tool errors, and ungrounded answers, and by a safety penalty $P_{\text{safety},i}$ for wrong answer-or-abstain decisions. With the floor of $0.80$ and the ceiling of $0.15$, every wrong rollout scores at least $0.65$ below every correct one.

\textbf{Advantage clamping.} GRPO centers each reward on its group mean, $A_i^{\text{raw}}=R(\tau_i)-\bar{R}$, without variance normalization~\cite{liu2025drgrpo}. When all rollouts in a group fail with different penalties, this centering still gives the least-penalized wrong rollout a positive advantage, a failure mode noted in recent analyses of shaped-reward GRPO~\cite{wang2026darkroom,yu2026filtermetric}. We set $A_i{=}0$ in all-wrong groups and in nearly flat groups whose reward range is below $0.15$, and we clamp the remaining wrong-rollout advantages to $A_i=\min(A_i^{\text{raw}},0)$. Wrong rollouts thus never receive a positive update ($\text{PAW}{=}0$), while informative all-right groups are retained to sharpen call efficiency. The policy is updated with a clipped token-level objective, turn-weighted assistant masking, and Schulman $k_3$ KL regularization~\cite{schulman2020klapprox,yu2025dapo}; exact coefficients are listed in Appendix~\ref{app:system_specs}.

% ==============================================================================
% Section IV: Experiments (V18 narrative rewrite: insight-first, MIMICEchoQA as closing note)
% ==============================================================================
\section{Experiments}
\label{sec:experiments}

% Table I (V14 Ultra-Clean Continuous-Tint tabularx): Zero Italic Sub-headers, Clean + Ours Hierarchy
\begin{table*}[t]
\centering
\caption{Main blind-test comparison on internal \bench{} (Study $N{=}242$, Single-Video $N{=}140$, Spatial $N{=}45$, Text $N{=}500$; zero missing predictions). Top block: tool-free direct video VLMs ($^\dagger$Answer Acc \%; Strict $\le 7.1\%$). Bottom block: tool-augmented base agents sharing identical frozen encoders (Concept-Grounded Strict Acc \%), followed by plug-and-play \sysname{} ($+$ Ours) and post-trained local $8\text{B}$ ($+$ Ours, trained); green parentheses show changes over each base agent.}
\label{tab:main_results}
\scriptsize
\setlength{\tabcolsep}{3.5pt}
\begin{tabularx}{\textwidth}{lCCCCCC}
\toprule
Method & Study & Single-Video & Spatial & Text & Study Refusal & Study Inv $\downarrow$ \\
\midrule
MedGemma-1.5-4B$^\dagger$~\cite{sellergren2026medgemma15} & 31.40 & 35.71 & 15.56 & --- & 0.0\% (0/32) & --- \\
LLaVA-Video-7B$^\dagger$~\cite{zhang2024llavavideo} & 30.17 & 30.71 & 6.67 & --- & 0.0\% (0/32) & --- \\
Qwen3-VL-8B$^\dagger$~\cite{bai2025qwen3vl} & 17.77 & 21.43 & 0.00 & --- & 28.1\% (9/32) & --- \\
Qwen3-VL-32B-FP8$^\dagger$~\cite{bai2025qwen3vl} & 21.07 & 22.86 & 8.89 & --- & 3.1\% (1/32) & --- \\
\midrule
Qwen3-8B & 56.61 & 52.14 & \textbf{82.22} & 73.00 & 68.8\% (22/32) & 0.56 \\
\rowcolor{oursrow}\quad $+$ Ours & 68.60\,\gain{+12.0} & 67.86\,\gain{+15.7} & \textbf{82.22} & 91.40\,\gain{+18.4} & \textbf{100\%} (32/32) & 0.13 \\
\rowcolor{oursrow}\quad $+$ Ours (trained)$^\ddagger$ & 70.25\,\gain{+13.6} & 65.71\,\gain{+13.6} & \textbf{82.22} & 92.20\,\gain{+19.2} & \textbf{100\%} (32/32) & \textbf{0.01} \\
\midrule
DeepSeek-V4.1-Flash & 64.05 & --- & --- & --- & \textbf{100\%} (32/32) & 0.13 \\
\rowcolor{oursrow}\quad $+$ Ours & 69.83\,\gain{+5.8} & --- & --- & --- & \textbf{100\%} (32/32) & 0.09 \\
\midrule
Qwen3.8-Flash & 63.64 & 60.71 & \textbf{82.22} & 99.00 & \textbf{100\%} (32/32) & 0.16 \\
\rowcolor{oursrow}\quad $+$ Ours & \textbf{70.66}\,\gain{+7.0} & \textbf{72.86}\,\gain{+12.2} & \textbf{82.22} & \textbf{99.20}\,\gain{+0.2} & \textbf{100\%} (32/32) & 0.11 \\
\bottomrule
\end{tabularx}

\vspace{3pt}
\noindent\parbox{\textwidth}{\scriptsize Bold: best per column. $^\dagger$Direct VLMs ($\le 8$ frames) lack $[e_X]$ handles and pixel tools (16-frame + coverage-prompt checks in Appendix Table~\ref{tab:app_overnight_extra}(c)). $^\ddagger$Fixed selected $8\text{B}$ operating point (zero Text-scope training trajectories); alternative SFT+Scalar operating point reaches $71.49\%$ Study Strict ($0.02$ Inv; Table~\ref{tab:ablations}(b)). Full CIs and external MIMICEchoQA audit appear in Appendix Tables~\ref{tab:app_full_blind}--\ref{tab:app_text_spatial_blind} and~\ref{tab:app_mimicechoqa}.}
\end{table*}

\subsection{Benchmarks, Implementation Details, and Baselines}
\label{subsec:exp_setup}
\textbf{Benchmark construction.} We build \bench{}, a four-scope evaluation suite of $76$ tasks, from PhysioNet-credentialed MIMIC-IV-Echo~\cite{gow2026mimicecho} and the official EchoNet splits~\cite{ouyang2020echonetdynamic,duffy2022echonetlvh}. The Study and Single-Video scopes share $63$ task definitions that span direct measurements, binary findings, severity grades, derived formulas, guideline reasoning, composite reasoning, and unsupported pathologies requiring abstention. Study items use full multi-view examinations, whereas Single-Video items use a single $16$-frame clip. The Spatial scope adds three pixel-level EchoNet tasks, namely LV segmentation, parasternal caliper localization, and four-class ejection-fraction grading, and the Text scope contains ten multi-step hemodynamic reasoning tasks built from non-imaging MIMIC patients. Open numeric questions are scored against structured MIMIC measurements within unit-aligned tolerance bands. Multiple-choice questions are derived from unambiguous report statements after filtering indeterminate phrasing, and their options are shuffled deterministically so that correct answers are balanced across letters. Guideline rules follow ASE/EACVI~\cite{lang2015chamber,baumgartner2017aortic,nagueh2016diastolic,zoghbi2017regurgitation,mukherjee2025rightheart,nagueh2025diastolicupdate} and AHA/ACC~\cite{heidenreich2022hf} standards together with documented reporting cutoffs. The blind test sets contain $242$ Study, $140$ Single-Video, $45$ Spatial, and $500$ Text items; Appendix~\ref{app:benchmark} gives the task catalog, splits, and tolerances. To assess \emph{Attr-Corr} independently of the agent, we further run standalone blind tests on EchoNet-Dynamic, EchoNet-LVH, and $60$ MIMIC clinical heads.

\textbf{Leakage control.} All splits are disjoint at the patient level, and imaging and text patients do not overlap. Memory priors, calibration caches, and the training trajectories of the local planner are built with five-fold out-of-fold splits inside the training patients. At inference, the agent sees only interfaces free of ground truth, enforced by a cooperative in-process Python audit hook (Appendix~\ref{app:system_specs}). Eight tasks, six imaging and two text, are withheld from all planner training trajectories to test unseen-task generalization, while their deterministic formulas, rules, and some hand-authored playbooks remain available (Appendix Table~\ref{tab:app_exposure}).

\textbf{Implementation.} All tool-augmented agents share the same frozen encoders~\cite{ouyang2020echonetdynamic,duffy2022echonetlvh,holste2025panecho,vukadinovic2025echoprime}. The local planner is Qwen3-8B~\cite{yang2025qwen3}, warm-started with bf16 LoRA of rank $16$ and then trained with GRPO using groups of six rollouts and a KL weight of $0.08$ selected on Study validation. Under the PhysioNet MIMIC-IV-Echo data use agreement, every planner, including DeepSeek-V4.1-Flash and Qwen3.8-Flash, ran on privately self-hosted endpoints with no transmission to external cloud APIs. Runtime was profiled on one NVIDIA RTX PRO~6000 Blackwell Max-Q GPU.

\textbf{Baselines and metrics.} We compare against four tool-free video VLMs, namely MedGemma-1.5-4B~\cite{sellergren2026medgemma15}, LLaVA-Video-7B~\cite{zhang2024llavavideo}, and Qwen3-VL at 8B and 32B~\cite{bai2025qwen3vl}, and against three base agents built on Qwen3-8B, DeepSeek-V4.1-Flash, and Qwen3.8-Flash that use the same tools without our modules. We report Answer accuracy, concept-grounded Strict accuracy, refusal rate on unsupported queries, and invalid tool calls per query. Patient-clustered bootstrap confidence intervals and exact McNemar tests appear in Appendix~\ref{app:full_results}.

\subsection{Main Results}
\label{subsec:exp_main}

% Table II (V14 Ultra-Clean Side-by-Side tabularx): Compact 6-Row Ablation + Clean 2-Col Regime Table, Zero Italic Sub-headers
\begin{table*}[t]
\centering
\caption{Core module ablation and online visual calibration regimes (Concept-Grounded Strict Acc \%). (a)~Stepwise component progression on Qwen3-8B Study blind test ($N{=}242$; $210$ answerable, $32$ unsupported $U_*$). (b)~Plug-in visual calibration comparing output-only scalar MLP (\emph{+Scalar}) against attribution-conditioned 2D \emph{ZeroRGB-Attr} (\emph{+Attr}).}
\label{tab:ablations}
\scriptsize
\setlength{\tabcolsep}{3.0pt}
\noindent
\begin{minipage}[t]{0.49\textwidth}
\vspace{0pt}
\textbf{(a) Module ablation on Qwen3-8B (Study test, $N{=}242$)}\\[3pt]
\begin{tabularx}{\linewidth}{lCCCC}
\toprule
Configuration & Strict & Answerable & Refusal & Inv $\downarrow$ \\
\midrule
Qwen3-8B & 56.61 & 54.76 & 68.8\% & 0.56 \\
\quad $+$ Harness & 60.74 & 54.76 & \textbf{100\%} & 0.61 \\
\quad $+$ Memory & 67.36 & 63.81 & 90.6\% & 0.16 \\
\rowcolor{oursrow}\quad $+$ Memory $+$ Harness & 68.60 & 63.81 & \textbf{100\%} & 0.14 \\
\quad $+$ SFT & \textbf{70.66} & \textbf{66.19} & \textbf{100\%} & 0.02 \\
\quad $+$ RL & 70.25 & 65.71 & \textbf{100\%} & \textbf{0.01} \\
\bottomrule
\end{tabularx}
\end{minipage}\hfill
\begin{minipage}[t]{0.49\textwidth}
\vspace{0pt}
\textbf{(b) Online visual calibration regimes (Strict Acc \%)}\\[3pt]
\begin{tabularx}{\linewidth}{llCCC}
\toprule
Split & Model & None & +Scalar & +Attr \\
\midrule
Study Val & DeepSeek & 67.14 & --- & \textbf{75.24}\,\gain{+8.1} \\
Study Val & Qwen3.8 & 68.10 & --- & \textbf{75.24}\,\gain{+7.1} \\
Study Val & 8B RL & 67.62 & 69.52 & \textbf{73.33}\,\gain{+5.7} \\
\midrule
SV Test & 8B RL & 59.29 & \textbf{67.14} & 65.71\,\gain{+6.4} \\
SV Test & Qwen3.8 & 69.29 & --- & \textbf{72.86}\,\gain{+3.6} \\
Study Test & 8B SFT / RL & 70.66 / 69.83 & \textbf{71.49 / 71.07} & 70.66 / 70.25 \\
\bottomrule
\end{tabularx}
\end{minipage}

\vspace{3pt}
\noindent\parbox{\textwidth}{\scriptsize Panel~(a): \emph{Answerable} denotes Strict Acc on the $210$ answerable items; $+$SFT and $+$RL keep memory and harness, and $+$RL is the final model of Table~\ref{tab:main_results}; memory sub-layer isolation (Priors vs.\ Playbooks) and same-backbone LLM-Verifier controls appear in Appendix Table~\ref{tab:app_overnight_extra}. Panel~(b): \emph{+Scalar} uses an output-only MLP; \emph{+Attr} adds attribution maps (green: $\Delta$ vs.\ \emph{None}; Study Val $N{=}210$, SV Test $N{=}140$, Study Test $N{=}242$).}
\end{table*}

Table~\ref{tab:main_results} compares paradigms rather than raw backbones, since direct VLMs and tool agents differ in frame budget and tool access.

\textbf{Direct VLMs.} Tool-free VLMs remain far below every tool-augmented agent: even under lenient Answer accuracy they reach $17.8$--$31.4\%$ on Study, and the weakest matches a text-only planner that never sees images ($17.4\%$). They also lack scope awareness: MedGemma answers nearly every unsupported query, whereas Qwen3-VL-8B over-abstains on answerable ones (Appendix Table~\ref{tab:app_overnight_extra}(c)).

\textbf{Plug-and-play transfer.} Attaching memory, harness, and calibration improves every base agent without touching its weights. The local $8\text{B}$ planner benefits most, gaining $12.0$, $15.7$, and $18.4$ points on Study, Single-Video, and Text, because it commits the execution errors that memory and verification target most often; on Text, formula execution rises from $55.6\%$ to $100\%$. The two frontier agents gain $5.8$ and $7.0$ points on Study; since they already refuse every unsupported query, this gain comes entirely from better answerable reasoning, confirming clinical competence rather than mere safety filtering. On Spatial, all tool agents tie at $82.2\%$, limited by segmentation and keypoint specialists rather than planning.

\textbf{Local policy learning.} Verification-guided training lifts the $8\text{B}$ planner to $70.25\%$ Study Strict accuracy, matching point estimates of frontier agents equipped with \sysname{}, while issuing roughly ten times fewer invalid calls. The gain reflects execution discipline: training removes redundant and malformed calls, trims Spatial tool calls by $42\%$ without any invalid call, and transfers to Text although no Text trajectory was used. Two trade-offs remain: Single-Video accuracy dips slightly below the plug-and-play setting, and warm-start with scalar calibration offers a marginally higher Study point ($71.49\%$) than our fixed final model.

\subsection{Ablation Study and Unseen Task Generalization}
\label{subsec:exp_core_pillars}
\begin{figure*}[t]
\centering
\includegraphics[width=\textwidth]{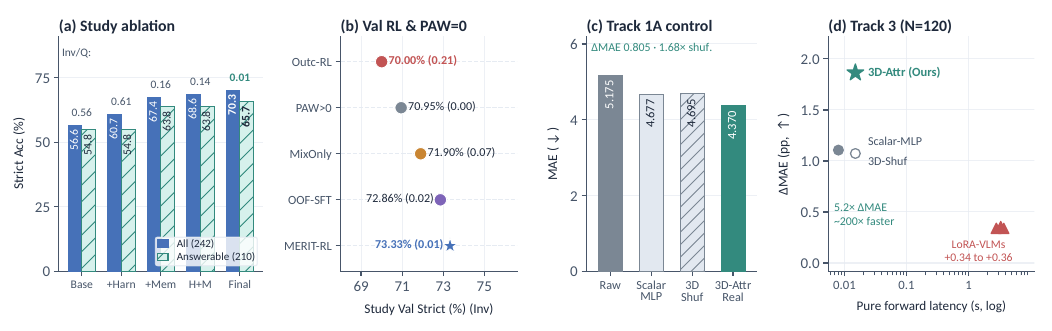}
\caption{Core mechanisms and plug-in visual calibration. (a)~Study blind-test progression separating overall Strict accuracy ($N{=}242$) from answerable Strict accuracy ($N{=}210$), with mean invalid calls (Inv/Q) above bars. (b)~Study Val post-training recipes ($N{=}210$; Strict Acc and Inv), which differ in initialization, reward, and curriculum; all RL arms share $100\%$ refusal, and the controlled three-seed comparison is in Appendix Table~\ref{tab:app_overnight_extra}(d). (c)~EchoNet-Dynamic Track~1A shuffled-heatmap control ($N{=}3{,}831$): $1.68\times$ the $\Delta\text{MAE}$ of shuffled heatmaps. (d)~EchoNet-Dynamic Track~3 paired blind test ($N{=}120$): $5.2\times$ larger $\Delta\text{MAE}$ at ${\sim}200\times$ faster pure forward latency than LoRA-VLMs.}
\label{fig:main_ablations}
\end{figure*}

Table~\ref{tab:ablations}(a) and Fig.~\ref{fig:main_ablations}(a)--(b) isolate each module on Qwen3-8B.

\textbf{Harness versus memory.} The two plug-and-play modules play complementary roles. The harness alone leaves answerable Study accuracy unchanged but closes the refusal gap completely, from $68.8\%$ to $100\%$, because unsupported queries are exactly what its scope gates detect. Memory alone produces the main jump on answerable queries, from $54.8\%$ to $63.8\%$, and cuts invalid calls by a factor of $3.5$, yet it still misses some unsupported queries. Separating the two memory layers (Appendix Table~\ref{tab:app_overnight_extra}(a)) reveals a clean division of labor: calibration priors alone raise answerable accuracy to $60.5\%$ while invalid calls barely move ($0.56\!\rightarrow\!0.51$), whereas workflow playbooks alone cut invalid calls to $0.16$. Combining harness and memory retains the strengths of both.

\textbf{Deterministic versus LLM verification.} Replacing the harness with a same-backbone LLM verifier leaves the refusal gap open ($90.6\%$ vs.\ $100\%$) and adds $41\%$ latency, while answerable accuracy stays statistically indistinguishable (Appendix Table~\ref{tab:app_overnight_extra}(b)). Deterministic checks thus deliver complete scope enforcement at lower cost, a guarantee that a stochastic reviewer does not provide.

% Table III (V14 Three Side-by-Side 4-Row Mini-Tables): Zero Italic Banners, No M1/M2 Tags, No Control Ratio, No Shading on Bold Rows
\begin{table*}[t]
\centering
\caption{Standalone plug-in visual calibration on EchoNet-Dynamic and EchoNet-LVH blind tests. (a)~EchoNet-Dynamic across three frozen encoders (Track~1A). (b)~Paired EchoNet-Dynamic comparison against LoRA-tuned video VLMs on EchoPrime (Track~3; pure forward time). (c)~EchoNet-LVH parasternal calipers (Track~1B, near ${\sim}0.16$\,cm/cell grid scale). Full per-encoder breakdowns, Track~2 MIMIC ($N{=}10{,}479$), and zero-shot VLMs appear in Appendix Table~\ref{tab:app_pred_extra}.}
\label{tab:error_predictor}
\scriptsize
\setlength{\tabcolsep}{2.4pt}
\noindent
\begin{minipage}[t]{0.32\textwidth}
\vspace{0pt}
\textbf{(a) EchoNet-Dynamic ($N{=}3{,}831$, pp)}\\[3pt]
\begin{tabularx}{\linewidth}{lCCC}
\toprule
Model & MAE$\downarrow$ & $\Delta$MAE$\uparrow$ & Harm$\downarrow$ \\
\midrule
Raw tools & 5.175 & --- & --- \\
Scalar-MLP & 4.677 & +0.498 & 44.1\% \\
3D-Shuf & 4.695 & +0.480 & 43.9\% \\
\textbf{3D-Attr (Ours)} & \textbf{4.370} & \textbf{\textcolor{gainc}{+0.805}} & \textbf{39.8\%} \\
\bottomrule
\end{tabularx}
\end{minipage}\hfill
\begin{minipage}[t]{0.345\textwidth}
\vspace{0pt}
\textbf{(b) Paired VLM Test ($N{=}120$, pp)}\\[3pt]
\begin{tabularx}{\linewidth}{lCCC}
\toprule
Model & MAE$\downarrow$ & $\Delta$MAE$\uparrow$ & Time \\
\midrule
Raw EchoPrime & 5.744 & --- & --- \\
LoRA-VLMs (3$\times$) & 5.39--5.40 & +0.34--0.36 & 2.8--3.7\,s \\
3D-Shuf & 4.670 & +1.073 & 15\,ms \\
\textbf{3D-Attr (Ours)} & \textbf{3.880} & \textbf{\textcolor{gainc}{+1.864}} & \textbf{15\,ms} \\
\bottomrule
\end{tabularx}
\end{minipage}\hfill
\begin{minipage}[t]{0.315\textwidth}
\vspace{0pt}
\textbf{(c) EchoNet-LVH ($N{=}1{,}357$, cm)}\\[3pt]
\begin{tabularx}{\linewidth}{lCCC}
\toprule
Model & MAE$\downarrow$ & $\Delta$MAE$\uparrow$ & Harm$\downarrow$ \\
\midrule
Raw tools & 0.244 & --- & --- \\
\textbf{Scalar-MLP} & \textbf{0.219} & \textbf{+0.024} & \textbf{44.0\%} \\
2D-Attr & 0.221 & +0.023 & 44.2\% \\
3D-Attr (Ours) & 0.221 & +0.023 & 44.7\% \\
\bottomrule
\end{tabularx}
\end{minipage}
\end{table*}

\textbf{Policy learning stability.} On Study validation (Fig.~\ref{fig:main_ablations}(b); Appendix Table~\ref{tab:posttrain}), warm-starting on our own trajectories performs at least as well as distillation from a frontier teacher. In the RL stage, a sparse outcome reward degrades the warm-start policy and inflates invalid calls about twentyfold, whereas our harness-guided reward improves on it ($73.3\%$ vs.\ $72.9\%$) with near-zero invalid calls. A controlled three-seed study (Appendix Table~\ref{tab:app_overnight_extra}(d)) confirms that advantage clamping removes every positive update on wrong rollouts and, together with cross-check credit, gives the highest validation mean ($73.0\%$). Blind Study accuracy is nearly identical across conditions ($70.3$--$70.4\%$, $\text{SD}\le 0.24$\,pp), so RL mainly consolidates execution discipline rather than adding accuracy.

\textbf{Unseen task generalization.} \sysname{} generalizes to the eight held-out tasks without any task-specific training trajectory (Appendix Table~\ref{tab:app_safety_latency}(b)). On the $28$ held-out Study items, Qwen3-8B rises from $53.6\%$ to $71.4\%$ with plug-and-play modules and to $78.6\%$ after training with scalar calibration, while both frontier agents rise from $64.3\%$ to $75.0\%$. The effect is largest on the $100$ held-out Text items, where the $8\text{B}$ planner climbs from $25.0\%$ to $66.0\%$ plug-and-play and $67.0\%$ after imaging-only training, and Qwen3.8 reaches $100\%$. Because held-out tasks reuse shared formulas, rules, and calibrated component heads, these gains indicate that the framework composes known procedures into new workflows instead of memorizing task-specific trajectories.

\textbf{Safety and runtime.} The full system refuses every unsupported test query autonomously, without forced truncation, while keeping false abstention below $1\%$ (Appendix~\ref{app:safety_runtime}). On five unseen ten-clip DICOM studies, an uncached first question takes $8.7$\,s end to end, dominated by video decoding and encoder passes; \emph{Attr-Corr} adds only about $18$\,ms, or $0.2\%$ of the total.

\subsection{Standalone Calibration and External Transfer}
\label{subsec:exp_error_predictor}
\textbf{Standalone calibration.} Table~\ref{tab:error_predictor} and Fig.~\ref{fig:main_ablations}(c)--(d) test whether attribution maps carry error signal beyond the raw prediction. On EchoNet-Dynamic, \emph{3D-Attr} lowers ejection-fraction MAE from $5.18$ to $4.37$ across three frozen encoders, whereas shuffling the attribution maps collapses the gain to the level of a scalar corrector. Real maps thus achieve $1.68\times$ the error reduction of shuffled ones, confirming that the correction depends on where the encoder looked rather than on output statistics alone. Against LoRA-tuned video VLMs on paired EchoPrime predictions, \emph{3D-Attr} achieves a $5.2\times$ larger error reduction at about $200\times$ lower latency, and zero-shot VLM critics even increase error (Appendix Table~\ref{tab:app_pred_extra}(c)). The benefit is bounded by attribution resolution. On EchoNet-LVH calipers, coarse feature maps of about $0.16$\,cm per cell leave little room for sub-millimeter refinement, so \emph{3D-Attr} only matches the scalar corrector, and the $60$-head MIMIC benchmark similarly splits wins between the two (Appendix Table~\ref{tab:app_pred_extra}(a)).

\textbf{Online calibration.} Table~\ref{tab:ablations}(b) shows that calibration helps most inside the agent when single-clip variance is high. On Study validation, 2D \emph{Attr-Corr} lifts both frontier agents to $75.2\%$ and improves every frozen $8\text{B}$ policy, whereas a LoRA-VLM critic lowers accuracy (Appendix Table~\ref{tab:app_hotswap}). On Single-Video blind tests, calibration improves every backbone by $3.6$ to $7.9$ points. On multi-view Study blind tests, pooling across views already smooths single-clip noise, so attribution calibration yields almost no net change and the lower-variance scalar corrector edges slightly ahead.

\textbf{External transfer.} We finally test zero-shot transfer on MIMICEchoQA~\cite{thapa2025openbiomedvid}, with all $582$ of its patients locked to testing (Appendix~\ref{app:external_audit}). Every \sysname{} backbone refuses all uncovered queries, and plug-and-play modules lift Qwen3.8 from $39.7\%$ to $46.1\%$. The trained $8\text{B}$ planner, however, falls below its base agent ($37.6\%$ vs.\ $49.0\%$), and 2D calibration also lowers accuracy. The benchmark's answer distribution differs sharply from ours: options are heavily skewed, so a blind policy that always selects option B scores $57.4\%$, and severity templates are triggered only by reported pathology. Training and calibration fitted to our balanced protocol appear to transfer poorly to such a distribution.

% ==============================================================================
% Section V: Conclusion and Limitations (V18.1)
% ==============================================================================
\section{Conclusion and Limitations}
\label{sec:conclusion}

\sysname{} combines read-only clinical experience, deterministic verification, and visual calibration for auditable echocardiography. Its modules improve frozen planners, while verification-guided training yields an $8\text{B}$ planner with comparable Study accuracy to frontier configurations and near-zero invalid calls. Attribution maps also improve standalone ejection-fraction calibration.

Several limitations remain. Attribution calibration is bounded by feature-map resolution on sub-millimeter calipers and adds little once multi-view pooling smooths clip-level noise. On an external benchmark with skewed answer priors, the trained planner and 2D calibration lose accuracy, calling for domain-adaptive training and calibration. Finally, evaluation relies on retrospective single-center cohorts and automated grading without prospective clinician adjudication. Task specifications, verification rules, and evaluation code will be released under PhysioNet credentialed access to support multi-center validation.

\bibliographystyle{IEEEtran}
\bibliography{IEEEabrv,reference}

% Structured appendix (Appendix A-E; tables/figures/equations numbered per appendix)
\clearpage
\useRomanappendicesfalse
\appendices
\renewcommand{\thesubsection}{\thesection.\arabic{subsection}}
\renewcommand{\thesubsectiondis}{\thesection.\arabic{subsection}}
\counterwithin*{table}{section}
\counterwithin*{figure}{section}
\counterwithin*{equation}{section}
\renewcommand{\thetable}{\thesection\arabic{table}}
\renewcommand{\thefigure}{\thesection\arabic{figure}}
\renewcommand{\theequation}{\thesection\arabic{equation}}
\renewcommand{\theHtable}{app.\thetable}
\renewcommand{\theHfigure}{app.\thefigure}
\renewcommand{\theHequation}{app.\theequation}
% ==============================================================================
% Supplementary Appendix (V18: Appendix A-E aligned one-to-one with the main text)
% ==============================================================================
\renewcommand{\topfraction}{0.95}
\renewcommand{\dbltopfraction}{0.95}
\renewcommand{\dblfloatpagefraction}{0.88}
\renewcommand{\textfraction}{0.05}

% ------------------------------------------------------------------------------
\section{Benchmark Construction, Task Catalog, and Held-Out Protocol}
\label{app:benchmark}
This appendix complements Section~\ref{subsec:exp_setup}. It lists all $76$ \bench{} tasks, describes the patient-level split of each scope, specifies the automated grading rules, and documents which resources remain available to the eight held-out tasks.

\begin{table*}[!t]
\centering
\caption{Complete \bench{} $76$-task catalog across four scopes and three complexity tiers, together with the held-out exposure matrix for the $6{+}2$ tasks ($^\star$) excluded from planner SFT and GRPO trajectories.}
\label{tab:app_exposure}
\scriptsize
\setlength{\tabcolsep}{2.4pt}
\begin{threeparttable}
\textbf{(a) Complete $76$-task catalog ($63$ Study/Single-Video + $3$ Spatial + $10$ Text; Tier~1: $48$, Tier~2: $14$, Tier~3: $14$ tasks; $^\star$held-out from planner SFT/RL)}\\[2pt]
\begin{tabular*}{\textwidth}{@{\extracolsep{\fill}}lc>{\raggedright\arraybackslash}p{0.66\textwidth}@{}}
\toprule
Category (Scope) & Count & Task IDs \\
\midrule
Direct Measurement (Study/SV) & 15 & \texttt{M\_EF}, \texttt{M\_IVSD}, \texttt{M\_LVIDD}, \texttt{M\_LVIDS}, \texttt{M\_LVPWD}, \texttt{M\_LVOT}, \texttt{M\_AOROOT}, \texttt{M\_LAIDS}, \texttt{M\_LAVOL}, \texttt{M\_LVEDV}, \texttt{M\_LVESV}, \texttt{M\_AV\_VMAX}, \texttt{M\_EE\_AVG}, \texttt{M\_TAPSE}, \texttt{M\_TR\_GRAD} \\
Binary Finding (Study/SV) & 15 & \texttt{B\_AOROOT\_DIL}, \texttt{B\_AR}, \texttt{B\_AS}, \texttt{B\_EFFUSION}, \texttt{B\_LA\_DIL}, \texttt{B\_LVH}, \texttt{B\_LV\_DIL}, \texttt{B\_MR}, \texttt{B\_PHTN}, \texttt{B\_RAP}, \texttt{B\_RA\_DIL}, \texttt{B\_RV\_DIL}, \texttt{B\_RV\_DYSF}, \texttt{B\_TR}, \texttt{B\_WMA} \\
Severity Grading (Study/SV) & 9 & \texttt{S\_AR}, \texttt{S\_AS}, \texttt{S\_LA\_DIL}, \texttt{S\_LVH}, \texttt{S\_MR}, \texttt{S\_PHTN}, \texttt{S\_RA\_DIL}, \texttt{S\_TR}, \texttt{S\_RV\_DIL}$^\star$ \\
Derived Formula (Study/SV) & 8 & \texttt{D\_AV\_GRAD}, \texttt{D\_FS}, \texttt{D\_LAVI}, \texttt{D\_LVIDD\_MM}, \texttt{D\_LVMASS}, \texttt{D\_LVMI}, \texttt{D\_SV}, \texttt{D\_RWT}$^\star$ \\
Guideline Reasoning (Study/SV) & 6 & \texttt{G\_AV\_VEL\_SEV}, \texttt{G\_LAVI\_ABN}, \texttt{G\_LAVI\_SEV}, \texttt{G\_LVEF\_CAT}, \texttt{G\_LV\_GEOM}, \texttt{G\_TAPSE\_ABN}$^\star$ \\
Composite Reasoning (Study/SV) & 4 & \texttt{C\_AV\_MV\_TV\_REG}, \texttt{C\_BIVENT\_FUNC}, \texttt{C\_BIVENT\_SIZE}, \texttt{C\_LEFT\_REMODEL}$^\star$ \\
Unsupported Refusal (Study/SV) & 6 & \texttt{U\_ASD}, \texttt{U\_MVP}, \texttt{U\_PULM\_VALVE}, \texttt{U\_THROMBUS}, \texttt{U\_COARCTATION}$^\star$, \texttt{U\_PVL}$^\star$ \\
Pixel-Level Spatial & 3 & \texttt{SP\_LV\_SEG} (LV Dice $\ge 0.80$), \texttt{SP\_PLAX\_LANDMARKS} (4-keypoint caliper), \texttt{SP\_DYN\_EF\_GRADE} (4-class A4C EF) \\
Clinical Text Reasoning & 10 & \texttt{T\_DIASTOLIC\_ASE2016}, \texttt{T\_LV\_GEOMETRY}, \texttt{T\_AS\_HEMO}, \texttt{T\_LAVI\_GRADE}, \texttt{T\_RVSP\_PHTN}, \texttt{T\_FS\_RWT\_MULTI}, \texttt{T\_STRUCTURE\_TRIAGE}, \texttt{T\_NO\_IMAGE\_GUARD}, \texttt{T\_CO\_CI}$^\star$, \texttt{T\_HF\_PHENOTYPE}$^\star$ \\
\bottomrule
\end{tabular*}

\vspace{2pt}
\textbf{(b) Held-out task separation matrix for the $6{+}2$ tasks ($^\star$) excluded from planner SFT and GRPO trajectories}\\[2pt]
\begin{tabular*}{\textwidth}{@{\extracolsep{\fill}}llccccl@{}}
\toprule
Task ID & Scope & SFT/GRPO Traj. & Formula / Rule & Playbook & Train Cal. / Pred. & Notes \\
\midrule
\texttt{D\_RWT} & Study/SV & \textbf{Excl.\ (0)} & \texttt{rwt} & Present & Comp.\ heads & \texttt{IVSd}/\texttt{LVPWd}/\texttt{LVIDd} calibrated on seen tasks \\
\texttt{G\_TAPSE\_ABN} & Study/SV & \textbf{Excl.\ (0)} & \texttt{tapse} & --- & \texttt{TAPSE} head & \texttt{TAPSE} calibrated on seen \texttt{M\_TAPSE} \\
\texttt{C\_LEFT\_REMODEL} & Study/SV & \textbf{Excl.\ (0)} & --- & Present & Comp.\ heads & Composite playbook over calibrated heads \\
\texttt{S\_RV\_DIL} & Study/SV & \textbf{Excl.\ (0)} & --- & --- & \texttt{CAL\_MAP} & \texttt{RVSize} cutoff fitted on train severity \\
\texttt{U\_COARCTATION}, \texttt{U\_PVL} & Study/SV & \textbf{Excl.\ (0)} & \texttt{list\_cap.} & Present & \texttt{no\_op} & Unsupported pathologies (Gate~1) \\
\texttt{T\_CO\_CI}, \texttt{T\_HF\_PHENOTYPE} & Text & \textbf{Excl.\ (0)} & \texttt{SV/CO/hf} & Present & N/A (Text) & Pure-text held-out reasoning tasks \\
\bottomrule
\end{tabular*}
\end{threeparttable}
\end{table*}

\subsection{Cohort Splits and Text Scenarios}
\label{app:cohorts}
All internal cohorts derive from a unified global patient-level partition of our local MIMIC-IV-Echo~\cite{gow2026mimicecho} patient table ($91{,}373$ patients; $426$ imaging patients and $621$ studies; zero imaging--text patient overlap), and all $582$ external MIMICEchoQA patients~\cite{thapa2025openbiomedvid} are locked exclusively to Test. The four scopes are split as follows.
\begin{itemize}
\item \textbf{Study ($63$ tasks):} $621$ multi-view studies ($426$ pts, $25{,}875$ QA); $470$ train studies ($332$ pts), $210$ balanced Val QA ($44$ pts), and $242$ balanced Test QA ($48$ pts from the $50$-pt, $84$-study, $3{,}421$-QA test pool).
\item \textbf{Single-Video ($\text{SV}_{w_0}$, $63$ tasks):} deterministic $16$-frame $w_0$ clips ($97.4$--$97.8\%$ exact \texttt{clip\_key}); $3{,}195$-QA train pool ($628$ curated train QA), $105$ Val QA ($67$ studies), and $140$ Test QA ($84$ studies, $50$ pts).
\item \textbf{Spatial ($3$ tasks):} apical 4-chamber LV segmentation, PLAX 4-keypoint localization, and balanced 4-class EF ($120/45/45$ train/val/test QA).
\item \textbf{Text ($10$ tasks):} $2{,}500/250/500$ train/val/test multi-step clinical reasoning QA built from real MIMIC hemodynamic measurements plus two structured clinical scenario templates (\texttt{IVC}$\rightarrow$\texttt{RAP} $3/15$\,mmHg and $2\times 2$ heart-failure hemodynamic phenotype).
\end{itemize}
Table~\ref{tab:app_exposure}(a) lists all $76$ tasks across three complexity tiers.

\subsection{Automated Grading Protocol and Tolerances}
\label{app:grading}
Scoring is fully automated from offline references, without clinician adjudication in this version. The three rule families below define correctness and grounding for every task type.
\begin{itemize}
\item \emph{Concept-Grounded Strict Accuracy \& Abstention:} For answerable queries (\texttt{expected\_decision=ANSWER}), \texttt{concept\_grounded} requires \texttt{decision=ANSWER}, at least one valid cited handle $[e_X]\in\mathcal{E}$, and at least one non-demographic cited record whose \texttt{concept} or \texttt{native\_head} intersects the task's relevant-concept allow-list; abstaining scores incorrect (\texttt{false\_abstention}). For unsupported queries (\texttt{expected\_decision=ABSTAIN}), both correctness and grounding hold iff \texttt{decision=ABSTAIN}.
\item \emph{Open-Numeric Tolerances:} Predicted units are aligned to reference units (incompatible units fail); items pass iff $|e|\le\text{tol}$ (soft score $\exp(-|e|/\text{tol})$). Study/SV direct measurements: EF $\pm 6.0\%$, LVIDd/LVIDs/LAIDs $\pm 0.5$\,cm, IVSd/LVPWd/LVOT $\pm 0.2$\,cm, TAPSE/AoRoot $\pm 0.4$\,cm, LAVol $\pm 16.0$\,mL, LVEDV $\pm 22.0$\,mL, LVESV $\pm 15.0$\,mL, AV $V_{\max}$ $\pm 0.45$\,m/s, TR gradient $\pm 8.0$\,mmHg, $E/e'$ $\pm 3.5$. Derived formulas: LAVI $\pm 8.5$\,mL/m$^2$, FS $\pm 6.5\%$, RWT $\pm 0.09$, LV mass $\pm 45.0$\,g, LVMI $\pm 24.0$\,g/m$^2$, AV gradient $\pm 8.0$\,mmHg, SV $\pm 18.0$\,mL, LVIDd (mm) $\pm 5.0$\,mm. Text formulas use $\pm 0.08$--$1.5$ in native units.
\item \emph{Spatial \& 4-Class EF Rules:} \texttt{SP\_LV\_SEG} passes at binary LV mask $\text{Dice}\ge 0.80$; \texttt{SP\_PLAX\_LANDMARKS} passes at orientation-invariant mean Euclidean error $\text{MPE}\le 20.0$\,px across valid calipers; \texttt{G\_LVEF\_CAT} and \texttt{SP\_DYN\_EF\_GRADE} use ASE/EACVI 2015 4-class LVEF cutoffs via \texttt{check\_guideline('lvef')} (Normal $\ge 52\%$ male / $\ge 54\%$ female, collapsing Hyperdynamic to Normal on \texttt{SP\_DYN\_EF\_GRADE}; Mildly reduced $41\%$--$<52/54\%$; Moderately reduced $30\%$--$<41\%$; Severely reduced $<30\%$).
\end{itemize}

\subsection{Held-Out Task Protocol}
\label{app:heldout}
Table~\ref{tab:app_exposure}(b) details the six imaging and two text tasks that are excluded from all planner SFT and GRPO trajectories. These tasks are not hidden from the whole system. Their deterministic formulas and guideline rules remain registered, several hand-authored playbooks cover them, component heads such as IVSd, LVPWd, LVIDd, and TAPSE are calibrated on seen tasks, and the RV-size cutoff used by the held-out severity task is fitted on training patients. The held-out split therefore tests whether the planner can compose available procedures into workflows it never practiced, rather than whether it can answer without any supporting knowledge.

\subsection{Data Governance and Self-Hosting}
\label{app:governance}
All experiments operated strictly under the PhysioNet MIMIC-IV-Echo~\cite{gow2026mimicecho} Data Use Agreement. Every evaluated planner, including local Qwen3-8B, the direct video VLMs, and the larger DeepSeek-V4.1-Flash and Qwen3.8-Flash baselines, ran exclusively on privately self-hosted on-premise inference endpoints, with no transmission to external third-party cloud APIs.

% ------------------------------------------------------------------------------
\section{System Specifications, Formulas, Gates, and Training Details}
\label{app:system_specs}
This appendix provides the implementation-level details omitted from Section~\ref{sec:method}: concrete tool interfaces, the rule and gate registries, memory retrieval scores, \emph{Attr-Corr} training objectives, and the exact post-training configuration and reward equations.

\begin{table*}[!t]
\centering
\caption{Registries of (a)~the $19$ clinical guideline rule engines, (b)~the $8$ pre-answer gate families separating hard refusal from bounded retries, and (c)~the mapping between main-text names and experimental arm IDs.}
\label{tab:app_rules_gates}
\scriptsize
\setlength{\tabcolsep}{1.8pt}
\begin{tabular*}{\textwidth}{@{\extracolsep{\fill}}ll@{\hspace{10pt}}ll@{\hspace{10pt}}ll@{}}
\toprule
\multicolumn{2}{@{}l}{\textbf{(a) 19 Clinical Rules (\texttt{check\_guideline})}} & \multicolumn{2}{l}{\textbf{(b) Pre-Answer Gates (Gates~1--8b)}} & \multicolumn{2}{l@{}}{\textbf{(c) Paper $\leftrightarrow$ Arm Map}} \\
\cmidrule(r){1-2}\cmidrule(lr){3-4}\cmidrule(l){5-6}
Rule Key & Provenance & Gate & Trigger \& Action & Main-Text Name & Code / Arm ID \\
\midrule
\texttt{lvef}, \texttt{lavi}, \texttt{lvidd} & ASE/EACVI 2015~\cite{lang2015chamber} & \textbf{Gate 1} & Unsupp.\ $\rightarrow$ \textbf{Hard Refusal}$^\dagger$ & Attr-Corr (umbrella) & \texttt{ZeroRGB-Attr} / \texttt{3D-Attr} \\
\texttt{wall\_thick.}, \texttt{lv\_geom.} & ASE/EACVI 2015~\cite{lang2015chamber} & \textbf{Gate 1b} & No img/nums $\rightarrow$ \textbf{Hard Ref.}$^\dagger$ & ZeroRGB-Attr / 3D-Attr & \texttt{m2\_dist\_gated} (2D) / \texttt{M2-3D} \\
\texttt{frac.\_short.}, \texttt{la\_diam}, \texttt{ra\_len.} & ASE/EACVI 2015~\cite{lang2015chamber} & Gate 2 & Premature abs.\ $\rightarrow$ retry$^\ast$ & Scalar-Cal / Scalar-MLP & \texttt{M0\_MLP} (output-only) \\
\texttt{ao\_root} & ASE 2015~\cite{lang2015chamber} / MIMIC & Gate 3 & Missing/invalid \texttt{[eX]} $\rightarrow$ retry$^\ast$ & 3D-Shuf & \texttt{M2-3D I5\_shuf} \\
\texttt{av\_pk\_vel}, \texttt{as\_hemodyn.} & EACVI/ASE 2017~\cite{baumgartner2017aortic} & Gate 4 & Missing \texttt{calc.} $\rightarrow$ retry$^\ast$ & Zero-Shot (H+M) & \texttt{ZS Full (None)} \\
\texttt{diastolic\_ase2016} & ASE 2016/2025~\cite{nagueh2016diastolic,nagueh2025diastolicupdate} & Gate 5 & Missing guide.\ $\rightarrow$ retry$^\ast$ & Joint-SFT / +Teacher-SFT & \texttt{E1} / \texttt{E2} (+168 DeepSeek) \\
\texttt{e\_e\_prime} & ASE 2016/2025~\cite{nagueh2016diastolic,nagueh2025diastolicupdate} & Gate 6 & Multi-struct.\ $\rightarrow$ retry$^\ast$ & \textbf{OOF-SFT} & \textbf{\texttt{E3}} (Img-Only, 0\% Text) \\
\texttt{tapse}, \texttt{rvsp}, \texttt{rv\_s\_vel}, \texttt{rv\_diam} & ASE Right Heart~\cite{mukherjee2025rightheart} & Gate 7 & MCQ format $\rightarrow$ retry$^\ast$ & Outcome-RL / Harness-RL & \texttt{F1} (0/1) / \texttt{F3} (\texttt{harness-v2}) \\
\texttt{tr\_gradient} & ASE / MIMIC & Gate 8a & Numeric guard $\rightarrow$ retry$^\ast$ & PAW$>$0-RL / MixOnly-RL & \texttt{F5} / \texttt{F6} (\texttt{--mixed-only}) \\
\texttt{hf\_phenotype} & AHA/ACC 2022~\cite{heidenreich2022hf} & Gate 8b & Scalar \& \texttt{cm}/\texttt{mm} $\rightarrow$ retry$^\ast$ & \textbf{EchoMERIT-RL} & \textbf{\texttt{F7 upd20}} (\texttt{v3\_cf}) \\
\bottomrule
\end{tabular*}
\vspace{1pt}
\\{\scriptsize $^{\dagger}$Hard refusal uses \texttt{FORCED\_ABSTAIN}. $^\ast$Ordinary Gates~2--8b run $\le 2$ retries then release via \texttt{PASS\_BUDGET\_LIMIT}.}
\end{table*}

\subsection{Tool Interfaces and Verification Rules}
\label{app:tools}
\sysname{} registers six typed tools: \texttt{predict\_echo} (PanEcho $40$ heads [$21$ reg, $10$ bin, $9$ mc3]; EchoPrime $21$ heads [$19$ retrieval + $2$ reg]), \texttt{run\_specialist} (EchoNet-Dynamic DeepLabV3~\cite{chen2017deeplabv3} $112\times 112$ LV mask, EchoNet-LVH PLAX keypoints, EchoNet-Dynamic R(2+1)D-18 EF, ED/ES phase), \texttt{inspect\_study}, \texttt{calculate} ($16$ deterministic physiological formulas), \texttt{check\_guideline} ($19$ clinical rule engines), and \texttt{list\_capabilities}. Evidence auditing checks five physiological consistency pairs (e.g., inter-model EF divergence $\ge 10$\,pp) and flags concept substitution. Table~\ref{tab:app_rules_gates} enumerates all $19$ guideline engines, all $8$ pre-answer gate families, and the academic-to-code arm map. Ordinary Gates~2--8b allow at most two retries and then release the draft with a logged budget event, whereas Gates~1 and~1b enforce forced abstention.

\subsection{Memory Construction and Retrieval}
\label{app:memory}
Calibration priors fit $t^*=\arg\max_t(0.65\,\text{Acc}+0.35\,\text{BalAcc})$ over a $91$-point $[5\%,95\%]$ grid separately for study ($42$ heads) and single-video ($29$ PanEcho, extensible to $40$ with EchoPrime), with borderline half-width $d=\operatorname{clip}(0.25\,\sigma_{\text{score}},0.03,0.08)$. Playbooks and lessons are authored from training-cohort error analyses (zero test inspection). Given query tokens $Q$, entry keywords $K$, and regex matches $|\mathcal{R}(q,m)|$, playbooks score
\begin{equation}
\label{eq:app_mem_score}
S_{\text{pb}}(q,m) = 3.0\,|\mathcal{R}(q,m)| + 1.5\,\frac{|Q\cap K|}{\max(1,\min(|Q|,|K|))},
\end{equation}
adding $+0.5\,\mathbb{I}[\text{scope match}]$ when $S_{\text{pb}}>0$; lessons score $S_{\text{les}}=3.0\,|\mathcal{R}(q,m)|+|Q\cap K|$. The memory card is compact by construction, but its length is not hard-capped by a tokenizer.

\subsection{Attr-Corr Backends and Training Objectives}
\label{app:attrcorr}
Under the unified \emph{Attr-Corr} umbrella, \emph{3D-Attr} uses R(2+1)D-18 + BiomedBERT FiLM over $10$-channel $(V,C,T,112,112)$ inputs (\texttt{I5\_proto\_phase\_seg}) for dynamic/spatial EF (Tracks~1A and~3), whereas \emph{ZeroRGB-Attr} uses a 5-fold ImageNet ResNet-18 ensemble over $5$-channel zero-RGB attribution inputs (\texttt{I1\_proto\_only}, anchor-residual heads) across $60$ MIMIC clinical heads (Track~2 and online Study/SV). The multi-task training criterion averages active task-type losses per batch:
\begin{itemize}
\item \emph{Regression:} $\mathcal{L}_{\text{reg}} = \text{SmoothL1}(\hat{z}, z) + \lambda_{\text{nll}}\,\text{NLL}_{\text{het}}(\hat{z}, z, \log\sigma) + \lambda_{\text{l2}}\,\overline{\Delta z^2} + \lambda_{\text{pw}}\,\mathcal{L}_{\text{rank}}(|\hat{z}_{\text{det}}-z|, \sigma)$, where $\text{NLL}_{\text{het}} = \frac{1}{2}e^{-2\log\sigma}(\hat{z}-z)^2+\log\sigma$ and $\mathcal{L}_{\text{rank}}$ is an optional pairwise softplus ranking term on \emph{post-correction} standardized residuals $|\hat{z}_{\text{det}}-z|$ active only when $\lambda_{\text{pw}}{>}0$.
\item \emph{Binary \& 3-Class Classification:} $\mathcal{L}_{\text{cls}} = \mathcal{L}_{\text{BCE/CE}}(\ell_{\text{corr}}, y) + \lambda_{\text{risk}}\,\text{BCE}(r_{\text{logit}}, \mathbb{I}[\hat{y}_{\text{raw}}\neq y]) + \lambda_{\text{l2}}\,\overline{\Delta\ell^2}$, where the risk target is strictly \emph{raw frozen-head error} ($\hat{y}_{\text{raw}}\neq y$) and residual logits receive L2 shrinkage.
\end{itemize}
Trainer defaults are $\lambda_{\text{nll}}{=}0.2$, $\lambda_{\text{risk}}{=}0.3$, $\lambda_{\text{l2}}{=}10^{-3}$, and $\lambda_{\text{pw}}{=}0.0$ (exact per-run weights follow individual training scripts; optional ranking is not active unless $\lambda_{\text{pw}}{>}0$). Ground-truth labels are visible strictly during offline training and OOF threshold calibration, never at inference. Five-fold MIMIC checkpoints use non-nested per-fold validation epoch selection; at inference, regression exports the 5-fold median ($\tau_h{=}0$) while classification exports 5-fold mean probabilities gated by OOF-calibrated per-task-type $\tau_{\text{type}}$, with \texttt{fold\_agree} computing residual-sign majority fraction (regression), median-class vote fraction (binary), or mean-probability argmax vote fraction (3-class). Online actors use ground-truth-free interfaces and whitelisted caches guarded by a cooperative Python \texttt{sys.addaudithook} (returning \texttt{no\_op} on offline cache misses).

\subsection{Post-Training Configuration and RL Equations}
\label{app:posttrain_eqs}
Pure-imaging OOF-SFT (\texttt{E3}) trains \texttt{bf16} LoRA ($r{=}16,\alpha{=}32$, dropout $0.05$, $1$ epoch, $\text{lr}{=}3\times 10^{-5}$, cosine schedule, max length $6{,}144$) on $1{,}475$ imaging trajectories ($80/20$ programmatic and $85/15$ rejection-sampled memory keep/drop; Joint-SFT \texttt{E1} adds $457$ text rows totaling $1{,}932$). For GRPO, let $\tilde{n}_{\text{gate}}=\min(n_{\text{gate}},2)$, $\tilde{n}_{\text{err}}=\min(n_{\text{err}},2)$, $c_{\text{ex}}=\min(\max(0,n_{\text{calls}}-5),4)$, and $\mathbb{I}_{\text{spot}}=\mathbb{I}[n_{\text{gate}}=0 \land n_{\text{err}}=0 \land \text{grounded}]$. First, the baseline \texttt{harness-v2} (\texttt{F2}/\texttt{F3}, initialized from \texttt{E1}) reward computes
\begin{equation}
\label{eq:app_reward_v2}
R_{\text{v2}}(\tau) =
\begin{cases}
\begin{aligned}
&\max\!\big(0.80,\; 1.0 + R_{\text{clean}}^{\text{v2}} \\[-2pt]
&\quad + R_{\text{risk}} - 0.04\,c_{\text{ex}}\big),
\end{aligned} & y_{\text{strict}}=1, \\[8pt]
\begin{aligned}
&\operatorname{clip}\!\big(0.15\,s_{\text{soft}} - P_{\text{clean}} - P_{\text{risk}} \\[-2pt]
&\;\;- P_{\text{safety}} - 0.04\,c_{\text{ex}},\,{-}0.60,\,0.15\big),
\end{aligned} & y_{\text{strict}}=0,
\end{cases}
\end{equation}
where $R_{\text{clean}}^{\text{v2}}=0.15\,\mathbb{I}_{\text{spot}}-(1-\mathbb{I}_{\text{spot}})(0.08\tilde{n}_{\text{gate}}+0.05\tilde{n}_{\text{err}})$, $R_{\text{risk}}=0.20\,\mathbb{I}[(h_{\text{risk}}\lor b_{\text{trig}})\land x_{\text{done}}]$, $P_{\text{clean}}=0.15\tilde{n}_{\text{gate}}+0.15\tilde{n}_{\text{err}}+0.10\,\mathbb{I}[\neg\text{grounded}\land\text{ANS}]$, $P_{\text{risk}}=0.15\,\mathbb{I}[h_{\text{risk}}\land\neg x_{\text{done}}]$, and $P_{\text{safety}}=0.25\,\mathbb{I}[s_{\text{fail}}]$.

Our final EchoMERIT-RL (\texttt{F7}, initialized from \texttt{E3}) uses documented run settings: pure-imaging dual-tool pool ($240$ Study $+ 80$ SV), $G{=}6$, $\text{lr}{=}1.5\times 10^{-6}$, $\beta{=}0.08$, $T{=}0.80$, $30$ target updates (\texttt{upd20} selected on Val), $2$ PPO epochs, turn weight $w_t{=}0.5$, max seq.\ $5{,}120$, 3-stage easy-to-hard curriculum, and mixed frontier replay (random seed defaults to $42$ in the trainer script; no separate per-run seed manifest is archived). Its exact group-coupled reward (\texttt{v3\_cf}) is
\begin{equation}
\label{eq:app_reward_v3}
R_{\text{v3}}(\tau_i) =
\begin{cases}
\begin{aligned}
&\max\!\big(0.80,\; 1.0 + R_{\text{clean},i}^{\text{v3}} \\[-2pt]
&\quad + R_{\text{short},i} + R_{\text{rescue},i}\big),
\end{aligned} & y_{\text{strict},i}=1, \\[8pt]
\begin{aligned}
&\operatorname{clip}\!\big(0.15\,\tilde{s}_{\text{soft},i} - P_{\text{clean},i} \\[-2pt]
&\quad - P_{\text{safety},i},\; -0.60,\; 0.15\big),
\end{aligned} & y_{\text{strict},i}=0,
\end{cases}
\end{equation}
where $\tilde{s}_{\text{soft},i}=\operatorname{clip}(s_{\text{soft},i},0,1)$, $R_{\text{clean},i}^{\text{v3}}=+0.15$ when $\mathbb{I}_{\text{spot},i}=1$ and $-(0.12\tilde{n}_{\text{gate},i}+0.15\tilde{n}_{\text{err},i})$ otherwise; $R_{\text{short},i}=\max(-0.16, 0.10-0.04\max(0,n_{\text{calls},i}-n_{\min}))$ if $\mathbb{I}_{\text{spot},i}=1$ ($n_{\min}$ is the minimum tool-call count among spotless strictly correct peers in the group) and $0$ otherwise; and $R_{\text{rescue},i}=+0.15$ iff rollout $i$ performs a clean cross-check ($\text{cross\_checks}_i>0 \land n_{\text{err},i}=0$) in a group containing \emph{at least one failing non-cross-check peer and zero passing non-cross-check peers}. On incorrect rollouts ($y_{\text{strict},i}=0$), $P_{\text{clean},i}$ and $P_{\text{safety},i}$ follow Eq.~\eqref{eq:app_reward_v2} with no cost or risk penalty.

With token importance ratio $r_{i,t}(\theta)=\exp(\operatorname{clip}(\log\pi_\theta(x_{i,t}\mid x_{i,<t})-\log\pi_{\theta_{\text{old}}}(x_{i,t}\mid x_{i,<t}),-5,5))$, asymmetric clipping $\tilde{r}_{i,t}(\theta)=\operatorname{clip}(r_{i,t}(\theta),1-\epsilon_{\text{low}},1+\epsilon_{\text{high}})$ ($\epsilon_{\text{low}}{=}0.20,\epsilon_{\text{high}}{=}0.28$), turn-weighted assistant mask $m_{i,t}\in\{0,0.5,1\}$, active set $\mathcal{V}=\{i : |A_i|>10^{-6}\}$, $Z=\sum_{i\in\mathcal{V}}\sum_t m_{i,t}$, eval-mode dropout recomputation, and grad-clip $1.0$, policy updates minimize
\begin{equation}
\label{eq:app_grpo_obj}
\begin{aligned}
\mathcal{L}(\theta) = -\frac{1}{Z}\sum_{i\in\mathcal{V}}\sum_t m_{i,t}\Big[&\min\!\big(r_{i,t}(\theta)A_i,\;\tilde{r}_{i,t}(\theta)A_i\big) \\[-2pt]
&- \beta\,\widehat{D}_{\text{KL}}^{(k3)}(x_{i,t})\Big],
\end{aligned}
\end{equation}
\begin{equation}
\label{eq:app_k3_kl}
\begin{aligned}
\widehat{D}_{\text{KL}}^{(k3)}(x_{i,t}) &= \exp(\Delta_{i,t}) - \Delta_{i,t} - 1, \\
\Delta_{i,t} &= \operatorname{clip}\!\Big(\log\frac{\pi_{\text{ref}}(x_{i,t}\mid x_{i,<t})}{\pi_\theta(x_{i,t}\mid x_{i,<t})},-10,10\Big).
\end{aligned}
\end{equation}

% ------------------------------------------------------------------------------
\section{Full Quantitative Results and Ablation Addenda}
\label{app:full_results}
This appendix reports the complete statistics behind Tables~\ref{tab:main_results}--\ref{tab:error_predictor} and Fig.~\ref{fig:main_ablations}. All $p$ values are nominal exact McNemar tests on paired QA items, and confidence intervals are patient-clustered bootstrap intervals unless stated otherwise. Table abbreviations: \textit{Ans} (Answer accuracy), \textit{ConcG} (concept-grounded rate), \textit{Inv} (invalid calls per query), \textit{MC3} (three-class severity), \textit{gNet} (net risk-gated correct flips), and \textit{AURC} (area under the risk--coverage curve).

\subsection{Four-Scope Blind Tests}
\label{app:blind_tests}
Tables~\ref{tab:app_full_blind} and~\ref{tab:app_text_spatial_blind} give single-arm breakdowns for all four blind test sets with zero missing predictions. Two observations complement Section~\ref{subsec:exp_main}. First, both frontier families improve significantly on Study once the plug-and-play modules are attached, even though their refusal is already complete. Second, every tool agent saturates at $82.22\%$ on Spatial, so post-training changes only call efficiency there.

\begin{table*}[!t]
\centering
\caption{Single-arm blind-test breakdown for Study ($N{=}242$, 48 patients) and Single-Video ($N{=}140$, 50 patients; zero missing predictions).}
\label{tab:app_full_blind}
\scriptsize
\setlength{\tabcolsep}{2.2pt}
\begin{threeparttable}
\textbf{(a) Study-242 full arm breakdown ($N{=}242$, 48 patients; Attr-Corr uses 5-fold ZeroRGB-Attr backend)}\\[2pt]
\begin{tabular*}{\textwidth}{@{\extracolsep{\fill}}lcccccl@{}}
\toprule
Arm & Ans (\%) & Strict (\%) & 95\% CI (Clustered) & ConcG (\%) & Calls / Inv & Key Nominal Exact McNemar $p$ \\
\midrule
Qwen3-8B No-Tools & 17.36 & 9.92 & [6.86, 13.24] & 9.9 & 0.00 / 0.00 & Lower bound \\
Qwen3-8B Base & 56.61 & 56.61 & [49.53, 63.59] & 93.4 & 3.10 / 0.56 & $U_*$ $68.8\%$ (10/32 halluc.) \\
+Harness & 61.16 & 60.74 & [54.13, 67.69] & 95.5 & 3.06 / 0.61 & vs.\ Base: $+4.13$\,pp ($p{=}0.03088$) \\
+Memory & 67.36 & 67.36 & [60.69, 73.33] & 96.7 & 2.18 / 0.16 & vs.\ Base: $+10.74$\,pp ($p{=}0.00011$); $U_*$ 29/32 \\
Zero-Shot (H+M) & 69.42 & 68.60 & [62.25, 74.64] & 97.1 & 2.26 / 0.14 & vs.\ Base: $+11.98$\,pp ($p{=}0.00003$) \\
Zero-Shot + Attr-Corr & 69.83 & 68.60 & [62.16, 75.22] & 97.1 & 2.12 / 0.13 & vs.\ ZS H+M: $0.00$\,pp (18:18, $p{=}1.00$) \\
OOF-SFT (Uncorr.) & 71.49 & 70.66 & [63.67, 76.78] & 98.8 & 1.78 / 0.01 & Inv $-93\%$ vs.\ ZS \\
\textbf{OOF-SFT + Scalar-Cal} & \textbf{71.90} & \textbf{71.49} & [65.61, 77.17] & \textbf{99.2} & 1.82 / 0.02 & Highest 8B point est.; Unseen-Ans $71.4\%$ \\
OOF-SFT + Attr-Corr & 71.49 & 70.66 & [63.96, 77.25] & 98.8 & 1.80 / 0.02 & vs.\ ZS+Attr: $+2.07$\,pp (6:1, $p{=}0.125$) \\
EchoMERIT-RL (Uncorr.) & 70.25 & 69.83 & [62.86, 75.81] & 98.8 & \textbf{1.77 / 0.00} & Zero invalid calls across 242 QA \\
EchoMERIT-RL + Scalar-Cal & 71.90 & 71.07 & [65.02, 76.98] & 98.8 & 1.78 / 0.01 & vs.\ None: $+1.24$\,pp (18:15, $p{=}0.728$) \\
EchoMERIT-RL + Attr-Corr & 71.07 & 70.25 & [63.41, 76.85] & 98.4 & 1.78 / 0.01 & Beats DS ($0.0275$) \& Qw3.8 Base ($0.0166$) \\
Outcome-RL + Attr-Corr & 69.01 & 68.60 & [61.57, 75.44] & 98.4 & 1.74 / 0.05 & $-1.65$\,pp vs.\ RL ($p{=}0.344$), ${\sim}5\times$ Inv \\
\midrule
DeepSeek Base & 64.05 & 64.05 & [56.62, 71.98] & 100.0 & 3.07 / 0.13 & --- \\
DeepSeek + H+M & 69.83 & 69.83 & [62.92, 76.41] & 100.0 & 2.29 / 0.06 & vs.\ Base: $+5.79$\,pp (21:7, $p{=}0.0125$) \\
DeepSeek + Attr-Corr & 69.83 & 69.83 & [62.91, 76.84] & 100.0 & 2.42 / 0.09 & vs.\ Base: $+5.79$\,pp (26:12, $p{=}0.0336$) \\
Qwen3.8 Base & 63.64 & 63.64 & [57.30, 70.30] & 100.0 & 3.10 / 0.16 & --- \\
Qwen3.8 + H+M & 70.25 & 70.25 & [63.52, 76.65] & 100.0 & 2.79 / 0.09 & vs.\ Base: $+6.61$\,pp (23:7, $p{=}0.0052$) \\
Qwen3.8 + Attr-Corr & 70.66 & 70.66 & [64.17, 77.46] & 100.0 & 2.69 / 0.11 & vs.\ Base: $+7.02$\,pp (28:11, $p{=}0.0095$) \\
\bottomrule
\end{tabular*}

\vspace{2pt}
\textbf{(b) SV-140 single-arm breakdown (dual-tool PE+EP + shared-$w_0$; $N{=}140$, 50 patients)}\\[2pt]
\begin{tabular*}{\textwidth}{@{\extracolsep{\fill}}lcccccl@{}}
\toprule
Arm & Ans (\%) & Strict (\%) & 95\% CI (Clustered) & ConcG (\%) & Calls / Inv & Key Nominal Exact McNemar $p$ \\
\midrule
Qwen3-8B No-Tools & 20.00 & 13.57 & [7.53, 19.18] & 13.6 & 0.00 / 0.00 & Lower bound \\
Qwen3-8B Base & 52.86 & 52.14 & [43.65, 60.31] & 94.3 & 3.06 / 0.52 & $U_*$ $70.0\%$ (6/20 halluc.) \\
Zero-Shot (H+M) & 64.29 & 62.86 & [53.06, 71.10] & 97.1 & 3.19 / 0.32 & vs.\ Base: $+10.71$\,pp (21:6, $p{=}0.0059$) \\
Zero-Shot + Attr-Corr & 67.86 & 67.86 & [57.60, 77.14] & 99.3 & 2.71 / 0.26 & vs.\ Base: $+15.71$\,pp (32:10, $p{=}0.00094$) \\
OOF-SFT (Uncorr.) & 60.71 & 60.71 & [52.31, 67.77] & 96.4 & 2.21 / 0.05 & --- \\
OOF-SFT + Attr-Corr & 67.14 & 67.14 & [57.94, 75.90] & 98.6 & 1.94 / 0.07 & vs.\ None: $+6.43$\,pp (20:11, $p{=}0.150$) \\
EchoMERIT-RL (Uncorr.) & 60.00 & 59.29 & [50.41, 66.92] & 95.7 & 2.14 / 0.05 & --- \\
EchoMERIT-RL + Scalar-Cal & 67.14 & 67.14 & [57.25, 75.97] & 98.6 & 1.97 / 0.07 & vs.\ None: $+7.86$\,pp (20:9, $p{=}0.0614$) \\
EchoMERIT-RL + Attr-Corr & 65.71 & 65.71 & [56.52, 74.51] & 98.6 & 1.87 / 0.05 & vs.\ None: $+6.43$\,pp (18:9, $p{=}0.122$) \\
Outcome-RL + Attr-Corr & 65.71 & 65.71 & [55.74, 75.74] & 99.3 & 1.69 / \textbf{0.01} & vs.\ RL+Attr: $0.00$\,pp (4:4, $p{=}1.00$) \\
\midrule
Qwen3.8 Base & 60.71 & 60.71 & [50.00, 70.12] & 100.0 & 3.19 / 0.08 & --- \\
Qwen3.8 + H+M & 69.29 & 69.29 & [60.77, 77.24] & 100.0 & 3.04 / 0.04 & vs.\ Base: $+8.57$\,pp (18:6, $p{=}0.0227$) \\
\textbf{Qwen3.8 + Attr-Corr} & \textbf{72.86} & \textbf{72.86} & [64.06, 81.65] & \textbf{100.0} & 2.91 / 0.06 & vs.\ H+M $p{=}0.383$; vs.\ Base $p{=}0.0060$ \\
\bottomrule
\end{tabular*}
\end{threeparttable}
\end{table*}

\begin{table*}[!t]
\centering
\caption{Blind-test arm breakdowns for Text ($N{=}500$, no Text-scope training trajectories) and Spatial ($N{=}45$; zero missing predictions).}
\label{tab:app_text_spatial_blind}
\scriptsize
\setlength{\tabcolsep}{2.4pt}
\begin{threeparttable}
\textbf{(a) Text-500 full 7-arm blind test ($N{=}500$; OOF-SFT and EchoMERIT-RL receive $0\%$ text post-training)}\\[2pt]
\begin{tabular*}{\textwidth}{@{\extracolsep{\fill}}lcccccl@{}}
\toprule
Arm & Ans (\%) & Strict (\%) & 95\% CI (Bootstrap) & ConcG (\%) & Calls / Inv & Nominal Exact McNemar $p$ \\
\midrule
Qwen3-8B No-Tools & 64.40 & 4.80 & [3.00, 6.80] & 4.8 & 0.00 / 0.00 & $-59.60$\,pp grounding gap \\
Qwen3-8B Base & 78.40 & 73.00 & [69.20, 76.80] & 91.6 & 3.60 / 1.86 & vs.\ No-Tools: $+68.20$\,pp ($p{<}10^{-5}$) \\
Zero-Shot (H+M) & 91.40 & 91.40 & [89.00, 93.80] & \textbf{100.0} & 2.65 / 0.17 & vs.\ Base: $+18.40$\,pp (96:4, $p{<}10^{-5}$) \\
OOF-SFT (0\% Text) & 92.00 & 92.00 & [89.60, 94.40] & \textbf{100.0} & 2.57 / \textbf{0.03} & vs.\ ZS H+M: $+0.60$\,pp (4:1, $p{=}0.375$, ns) \\
EchoMERIT-RL (0\% Text) & 92.20 & 92.20 & [89.80, 94.60] & \textbf{100.0} & \textbf{2.54 / 0.03} & vs.\ ZS H+M: $+0.80$\,pp (5:1, $p{=}0.219$, ns) \\
\midrule
Qwen3.8 Base & 99.00 & 99.00 & [98.00, 99.80] & \textbf{100.0} & 3.05 / 0.24 & --- \\
\textbf{Qwen3.8 + H+M} & \textbf{99.20} & \textbf{99.20} & [98.40, 99.80] & \textbf{100.0} & 2.78 / 0.07 & vs.\ Base: $+0.20$\,pp (3:2, $p{=}1.00$, ns) \\
\bottomrule
\end{tabular*}

\vspace{2pt}
\textbf{(b) Spatial-45 full 8-arm tool-agent and 4-VLM blind test ($N{=}45$; 3D-Attr used on EF clip task)}\\[2pt]
\begin{tabular*}{\textwidth}{@{\extracolsep{\fill}}lcccccc@{}}
\toprule
Arm & Ans (\%) & Strict (\%) & 95\% CI (Bootstrap) & EF-15 Ans (\%) & ConcG (\%) & Calls / Inv \\
\midrule
MedGemma-1.5-4B (Direct VLM)$^\star$ & 15.56 & 0.00 & [0.00, 0.00] & 46.7 & 0.0 & --- \\
LLaVA-Video-7B (Direct VLM)$^\star$ & 6.67 & 0.00 & [0.00, 0.00] & 20.0 & 0.0 & --- \\
Qwen3-VL-8B (Direct VLM)$^\star$ & 0.00 & 0.00 & [0.00, 0.00] & 0.0 & 0.0 & --- \\
Qwen3-VL-32B-FP8 (Direct VLM)$^\star$ & 8.89 & 0.00 & [0.00, 0.00] & 26.7 & 0.0 & --- \\
\midrule
Qwen3-8B No-Tools & 0.00 & 0.00 & [0.00, 0.00] & 0.0 & 0.0 & 0.00 / 0.00 \\
Qwen3-8B Base & 82.22 & 82.22 & [71.11, 93.33] & 80.0 & 100.0 & 2.31 / 1.20 \\
Zero-Shot (None / +3D-Attr) & 82.22 & 82.22 & [71.11, 93.33] & 80.0 & 100.0 & 2.69 / 0.91 \\
OOF-SFT (None / +3D-Attr) & 82.22 & 82.22 & [71.11, 93.33] & 80.0 & 100.0 & \textbf{1.33 / 0.00} \\
\textbf{EchoMERIT-RL (None / +3D-Attr)} & \textbf{82.22} & \textbf{82.22} & [71.11, 93.33] & \textbf{80.0} & \textbf{100.0} & \textbf{1.33 / 0.00} \\
\bottomrule
\end{tabular*}
\begin{tablenotes}[flushleft]
\scriptsize
\item Panel~(b): Tool agents score $93.3\%$ LV seg ($14/15$), $73.3\%$ PLAX keypoints ($11/15$), and $80.0\%$ EF ($12/15$); None and +3D-Attr match identically ($p{=}1.00$). $^\star$Direct VLM adapter abstains on the 30 segmentation/landmark items and scores the 15 EF clips.
\end{tablenotes}
\end{threeparttable}
\end{table*}

\subsection{Validation Ablations, Hot-Swap Transfer, and Post-Training Sweeps}
\label{app:val_ablations}
Table~\ref{tab:app_hotswap} reports component ablations on the Text, Single-Video, and Spatial validation splits, memory-size scaling, and a hot-swap matrix that pairs each frozen policy with each calibration backend on Study validation. Table~\ref{tab:posttrain} compares post-training recipes on locked validation splits, and Table~\ref{tab:app_posttrain_ext} adds the initialization lineage, single-video interface, and checkpoint sweep of every arm. Arms with different single-video interfaces are never compared on combined totals.

\begin{table*}[!t]
\centering
\caption{Validation component ablations (Text Val~250, SV Val~105, Spatial Val~45), memory-size scaling (B2), two-frontier Study Val~210 transfers (DeepSeek and Qwen3.8), and frozen-policy hot-swap matrix on Study Val~210 (zero missing predictions).}
\label{tab:app_hotswap}
\scriptsize
\setlength{\tabcolsep}{2.2pt}
\begin{threeparttable}
\textbf{(a) Validation component ablations on Text, Single-Video, and Spatial scopes (Qwen3-8B; single-arm rows)}\\[2pt]
\begin{tabular*}{\textwidth}{@{\extracolsep{\fill}}llcccccl@{}}
\toprule
Scope & Arm & Ans & Strict & 95\% CI & ConcG & Calls/Inv & Nominal Exact McNemar $p$ \\
\midrule
\multirow{5}{*}{Text Val 250}
& No-Tools & 59.20 & 4.80 & [2.40, 7.60] & 4.8 & 0.00/0.00 & $-54.40$\,pp grounding gap \\
& Base & 77.60 & 74.00 & [68.80, 79.20] & 94.4 & 3.67/1.91 & --- \\
& +Harness & 79.60 & 79.20 & [74.00, 84.00] & 98.4 & 3.23/1.50 & $0.00720$ (17:4, Base$\rightarrow$+H) \\
& +Memory & 90.80 & \textbf{90.80} & [87.20, 94.40] & 100.0 & 2.67/0.23 & $<10^{-5}$ (46:4, Base$\rightarrow$+M) \\
& Zero-Shot (H+M) & 90.00 & 90.00 & [86.40, 93.60] & 100.0 & 2.65/0.18 & $<10^{-5}$ (45:5, Base$\rightarrow$H+M) \\
\midrule
\multirow{4}{*}{SV Val 105}
& No-Tools & 14.29 & 12.38 & [7.08, 18.00] & 12.4 & 0.00/0.00 & --- \\
& Base & 47.62 & 45.71 & [36.00, 54.46] & 89.5 & 3.23/0.60 & --- \\
& Zero-Shot (H+M) & 57.14 & 57.14 & [47.31, 66.98] & 98.1 & 3.23/0.33 & $0.01690$ (17:5, Base$\rightarrow$H+M) \\
& Zero-Shot + Attr-Corr & 60.95 & \textbf{60.95} & [50.45, 71.30] & 99.1 & 2.70/0.30 & $0.00700$ (24:8, Base$\rightarrow$+Attr) \\
\midrule
\multirow{4}{*}{Spatial Val 45}
& No-Tools & 0.00 & 0.00 & [0.00, 0.00] & 0.0 & 0.00/0.00 & --- \\
& Base & 75.56 & 75.56 & [62.22, 86.67] & 100.0 & 2.31/1.40 & --- \\
& Zero-Shot (None) & 75.56 & 75.56 & [62.22, 86.67] & 100.0 & 2.64/0.96 & --- \\
& Zero-Shot + 3D-Attr & 77.78 & \textbf{77.78} & [64.44, 88.89] & 100.0 & 2.51/0.89 & $1.00$ vs.\ None (EF $53\!\rightarrow\!60\%$) \\
\bottomrule
\end{tabular*}

\vspace{2pt}
\textbf{(b) Study Val 210 hot-swap matrix, memory-size scaling (B2), and two-frontier transfers (grouped rows report intermediate CIs in note)}\\[2pt]
\begin{tabular*}{\textwidth}{@{\extracolsep{\fill}}llcccccc@{}}
\toprule
Policy & Backend & Ans & Strict & 95\% CI & Calls/Inv & $\Delta$ vs.\ None & Nominal Exact $p$ \\
\midrule
ZS (B2 Mem) & 25\% / 50\% / 100\% & 65.24 / 64.29 / 66.19 & 63.81 / 62.86 / 64.76 & [58.33, 71.37]$^\dagger$ & 2.30/0.19$^\dagger$ & --- & --- \\
\midrule
\multirow{5}{*}{ZS (H+M)}
& None & 66.19 & 64.76 & [58.33, 71.37] & 2.30/0.19 & --- & --- \\
& VLM (LLaVA-7B) & 62.38 & 61.43 & [54.90, 67.83] & 2.31/0.15 & $-3.33$\,pp & --- \\
& Scalar-Cal & 66.19 & 65.24 & [58.22, 72.73] & 2.12/0.11 & $+0.48$\,pp & --- \\
& M2-plain (ungated) & 67.62 & 67.14 & [60.34, 74.23] & 2.14/0.09 & $+2.38$\,pp & --- \\
& Attr-Corr (ZeroRGB) & 68.10 & \textbf{67.14} & [61.28, 73.97] & 2.11/0.11 & $+2.38$\,pp & $0.48685$ (ns) \\
\midrule
OOF-SFT & None / Scalar / Attr & 68.10 / 69.52 / \textbf{72.86} & 68.10 / 69.52 / \textbf{72.86} & [67.01, 79.15]$^\dagger$ & 1.81/0.02$^\dagger$ & $+1.43$/$+4.76$ & $0.07552$ (18:8) \\
Harness-RL & None / Scalar / Attr & 65.71 / 69.05 / \textbf{71.90} & 65.71 / 69.05 / \textbf{71.90} & [65.50, 78.31]$^\dagger$ & 1.74/0.00$^\dagger$ & $+3.33$/$+6.19$ & $0.04096$ (24:11) \\
EchoMERIT-RL & None / Scalar / Attr & 67.62 / 69.52 / \textbf{73.33} & 67.62 / 69.52 / \textbf{73.33} & [67.72, 79.23]$^\dagger$ & 1.80/0.01$^\dagger$ & $+1.90$/$+5.71$ & $0.03570$ (20:8) \\
\midrule
DeepSeek & Base / +H+M / +Attr & 66.19 / 67.14 / \textbf{75.24} & 66.19 / 67.14 / \textbf{75.24} & [69.32, 81.48]$^\dagger$ & 2.64/0.03$^\dagger$ & $+8.10$\,pp & \textbf{0.00455} (25:8) \\
Qwen3.8 & Base / +H+M / +Attr & 67.62 / 68.10 / \textbf{75.24} & 67.62 / 68.10 / \textbf{75.24} & [69.63, 81.03]$^\dagger$ & 2.74/0.02$^\dagger$ & $+7.14$\,pp & \textbf{0.00813} (22:7) \\
\bottomrule
\end{tabular*}
\begin{tablenotes}[flushleft]
\scriptsize
\item $^{\dagger}$CI and Calls/Inv shown for rightmost (+Attr / 100\%) arm; intermediate CIs: DeepSeek Base $[59.90,72.47]$, +H+M $[60.33,73.63]$; Qwen3.8 Base $[61.43,74.09]$, +H+M $[61.33,74.27]$; OOF-SFT None $[61.33,74.53]$, Scalar $[63.84,75.38]$; Harness-RL None $[59.07,72.20]$, Scalar $[63.35,75.11]$; EchoMERIT-RL None $[60.73,74.15]$, Scalar $[63.87,75.38]$.
\end{tablenotes}
\end{threeparttable}
\end{table*}

% Appendix Table (V13 Moved to Appendix with B2.1 Completed Text Val Cells): Post-Training Recipe Comparison
\begin{table*}[!t]
\centering
\caption{Post-training recipe comparison on locked validation splits (Study Val $N{=}210$, Text Val $N{=}250$; all using \emph{Attr-Corr}, zero missing predictions; visual summary in main Fig.~\ref{fig:main_ablations}(b)). Stage~2 recipes differ in initialization, reward, group/KL settings, and curriculum/replay; strictly controlled 3-seed $\text{PAW}=0$ swaps appear in Table~\ref{tab:app_overnight_extra}(d).}
\label{tab:posttrain}
\scriptsize
\setlength{\tabcolsep}{2.6pt}
\begin{threeparttable}
\begin{tabular*}{\textwidth}{@{\extracolsep{\fill}}llcccl@{}}
\toprule
Training Recipe & Init. & Study Val (Ans/Strict) & Text Val (\%) & Inv $\downarrow$ & $\Delta$ vs.\ Init.\ / Key Comparison \\
\midrule
\rowcolor{sectbg}\multicolumn{6}{@{}l}{\textit{Stage 1: Patient-OOF SFT (1 epoch; no external LLM teacher distillation except +Teacher-SFT)}} \\
Zero-Shot + Attr-Corr & Base & 68.10 / 67.14 & 90.00 & 0.11 & Reference without post-training \\
Joint-SFT (Image + Text, no teacher) & Base & 71.43 & 90.40 & 0.01 & $+4.29$\,pp vs.\ ZS; ${\sim}11\times$ lower Inv \\
+Teacher-SFT (+168 DeepSeek traj.) & Base & 70.95 & 90.40 & 0.03 & $+3.81$\,pp vs.\ ZS ($p{=}1.00$ vs.\ Joint) \\
\rowcolor{oursrow}\textbf{OOF-SFT (Img-only, no teacher)} & Base & \textbf{72.86} & \textbf{90.40} & \textbf{0.02} & $+5.71$\,pp vs.\ ZS ($p{=}0.0005$) \\
\midrule
\rowcolor{sectbg}\multicolumn{6}{@{}l}{\textit{Stage 2: Multi-Turn RL (single-run recipe comparison; Text Val completed with zero missing predictions)}} \\
Outcome-RL (sparse 0/1 reward) & Joint-SFT & 70.48 / 70.00 & 89.60 & 0.21 & $-1.43$\,pp vs.\ init; Inv $0.01\!\rightarrow\!0.21$ (${\sim}21\times$) \\
PAW$>$0-RL (harness reward, no guard) & OOF-SFT & 70.95 & 90.00 & 0.00 & $-1.90$\,pp vs.\ init \\
MixOnly-RL (mixed-group-only) & OOF-SFT & 71.90 & \textbf{95.20} & 0.07 & $-0.95$\,pp Study Val vs.\ init \\
\rowcolor{bestrow}\textbf{EchoMERIT-RL (PAW$=$0 + cross-check)} & OOF-SFT & \textbf{73.33} & 90.40 & \textbf{0.01} & $+0.48$\,pp vs.\ init ($+6.19$\,pp vs.\ ZS, $p{=}0.0002$) \\
\bottomrule
\end{tabular*}
\end{threeparttable}
\end{table*}

\begin{table*}[!t]
\centering
\caption{Four-scope post-training validation matrix with explicit initialization lineage, single-video interface separation (\texttt{PE+EP} vs.\ archived \texttt{PE-only}), and EchoMERIT-RL (\texttt{F7}) checkpoint sweep.}
\label{tab:app_posttrain_ext}
\scriptsize
\setlength{\tabcolsep}{2.1pt}
\begin{threeparttable}
\begin{tabular*}{\textwidth}{@{\extracolsep{\fill}}lllcccccc@{}}
\toprule
Arm (Code ID) & Init. & SV Intf. & Study 210 [95\% CI] & SV $w_0$ 105 & Spatial 45 & Text 250 [95\% CI] & ConcG & Study Calls / Inv \\
\midrule
Joint-SFT (E1) & Base & PE-only & 71.43 [65.5, 77.6] & 61.90/60.95 & 77.78 & 90.40 [86.8, 94.0] & 99.5 & 1.79 / 0.01 \\
+Teacher-SFT (E2) & Base & PE-only & 70.95 [64.6, 77.5] & 61.90 & 77.78 & 90.40 [86.8, 94.0] & 99.5 & 1.86 / 0.03 \\
\textbf{OOF-SFT (E3)} & Base & \textbf{PE+EP} & \textbf{72.86 [67.0, 79.2]} & \textbf{62.86} & 77.78 & 90.40 [86.8, 94.0] & 99.5 & 1.81 / 0.02 \\
\midrule
Outcome-RL (F1) & Joint-SFT & PE-only & 70.48/70.00 [63.8, 76.5] & 59.05 & --- & --- & 98.6 & 1.76 / 0.21 \\
Harness-v2 (F2) & Joint-SFT & PE-only & 70.95 [65.0, 77.2] & 61.90 & 77.78 & 90.00 [86.4, 93.6] & 100.0 & 1.67 / 0.00 \\
Harness-RL (F3) & Joint-SFT & PE-only & 71.90 [65.5, 78.3] & 62.86 & 77.78 & 90.40 [86.8, 94.0] & 99.5 & 1.74 / 0.00 \\
Hard-LowKL (F4) & Joint-SFT & PE-only & 70.95 [64.7, 77.4] & 63.81 & 77.78 & 90.80 [87.2, 94.4] & 100.0 & 1.77 / 0.00 \\
PAW$>$0-RL (F5) & OOF-SFT & PE-only & 70.95 [64.9, 77.4] & 60.95 & 77.78 & --- & 99.5 & 1.69 / 0.00 \\
MixOnly-RL (F6) & OOF-SFT & PE-only & 71.90 [65.7, 78.3] & 59.05/58.10 & --- & --- & 100.0 & 1.93 / 0.07 \\
\midrule
EchoMERIT-RL upd5 (F7) & OOF-SFT & PE+EP & not archived & 60.00 & --- & --- & --- & --- (SV Inv 0.05) \\
EchoMERIT-RL upd10 (F7) & OOF-SFT & PE+EP & 72.86 [67.0, 79.1] & 60.95 & --- & --- & 100.0 & 1.77 / 0.00 \\
EchoMERIT-RL upd15 (F7) & OOF-SFT & PE+EP & 72.86 [67.3, 78.7] & 61.90 & 77.78 & 90.40 [86.8, 94.0] & 100.0 & 1.75 / 0.01 \\
\textbf{EchoMERIT-RL upd20 (F7)} & OOF-SFT & \textbf{PE+EP} & \textbf{73.33 [67.7, 79.2]} & 59.05 & 77.78 & 90.40 [86.8, 94.0] & \textbf{100.0} & 1.80 / 0.01 \\
EchoMERIT-RL upd25 (F7) & OOF-SFT & PE+EP & not archived & 60.00 & --- & --- & --- & --- (SV Inv 0.03) \\
EchoMERIT-RL final (F7) & OOF-SFT & PE+EP & 72.86 [67.3, 78.7] & 58.10 & --- & --- & 100.0 & 1.79 / 0.01 \\
\bottomrule
\end{tabular*}
\begin{tablenotes}[flushleft]
\scriptsize
\item Arms using dual-tool \texttt{PE+EP} SV are never cross-compared on combined totals against archived \texttt{PE-only} SV arms. \texttt{F1}--\texttt{F4} initialize from Joint-SFT (\texttt{E1}); \texttt{F5}--\texttt{F7} initialize from OOF-SFT (\texttt{E3}). Checkpoints \texttt{upd5}/\texttt{upd25} were evaluated on SV-105 ($60.00\%$ each; Inv $0.05$/$0.03$) rather than Study-210.
\end{tablenotes}
\end{threeparttable}
\end{table*}

\subsection{Standalone Calibration Addenda}
\label{app:calib_addenda}
Table~\ref{tab:app_pred_extra} completes the standalone calibration benchmarks of Table~\ref{tab:error_predictor} with the $60$-head MIMIC evaluation, the full EchoNet-LVH breakdown, and zero-shot VLM diagnostics. The zero-shot VLM rows come from the EchoPrime validation split and are therefore reported separately from the paired blind test in the main text.

\begin{table*}[!t]
\centering
\caption{Supplementary standalone error-predictor evaluations: (a)~Track~2 MIMIC 60-task 11-metric blind test ($N{=}10{,}479$, 50 pts / 84 studies), (b)~Track~1B EchoNet-LVH PLAX blind test ($N{=}1{,}357$, cm), and (c)~Track~3 validation zero-shot VLM diagnostics ($N{=}120$ Val).}
\label{tab:app_pred_extra}
\scriptsize
\setlength{\tabcolsep}{2.2pt}
\begin{threeparttable}
\textbf{(a) Track~2: MIMIC 60-task 5-fold blind test ($N{=}10{,}479$ prediction rows, 50 pts / 84 studies)}\\[2pt]
\begin{tabular*}{\textwidth}{@{\extracolsep{\fill}}lccccccccccc@{}}
\toprule
Arm & Reg MAE & Harm (\%) & AURC & Bin AUROC & Bin F1 & Bin ECE & Bin gNet & MC3 AUROC & MC3 F1 & MC3 ECE & MC3 gNet \\
\midrule
Without correction & 5.3834 & --- & --- & 0.6772 & 0.2420 & 0.2883 & 0 & 0.6342 & 0.2909 & 0.3242 & 0 \\
5-Fold Scalar-MLP (M0) & \textbf{5.0553} & 41.25 & 3.8671 & 0.6940 & \textbf{0.4325} & \textbf{0.1775} & $+256$ & \textbf{0.6648} & 0.4076 & 0.1069 & \textbf{$+120$} \\
5-Fold ZeroRGB-Attr (M2, ResNet-18) & 5.0811 & \textbf{40.74} & \textbf{3.8493} & \textbf{0.6952} & 0.4278 & 0.1793 & \textbf{$+262$} & 0.6548 & \textbf{0.4190} & \textbf{0.1025} & $+101$ \\
\bottomrule
\end{tabular*}

\vspace{2pt}
\textbf{(b) Track~1B: EchoNet-LVH PLAX blind test ($N{=}1{,}357$; raw macro MAE $0.2436$\,cm; ${\sim}0.16$\,cm/cell grid scale)}\\[2pt]
\begin{tabular*}{\textwidth}{@{\extracolsep{\fill}}lccccc@{}}
\toprule
Model & Macro MAE (cm) & $\Delta$MAE (cm) & Harm (\%) & AURC & LVPWd MAE (cm) \\
\midrule
Raw frozen tools & 0.2436 & --- & --- & --- & 0.2225 \\
\textbf{Scalar-MLP (M0)} & \textbf{0.2192} & $+0.0244$ & 44.01 & 0.1910 & 0.1468 \\
2D-Attr (M2-2D I1) & 0.2210 & $+0.0226$ & 44.22 & 0.1992 & 0.1497 \\
3D-Attr (M2-3D I4) & 0.2208 & $+0.0228$ & 44.74 & 0.1935 & \textbf{0.1456} \\
\bottomrule
\end{tabular*}

\vspace{2pt}
\textbf{(c) Track~3 validation zero-shot VLM diagnostics ($N{=}120$ EchoPrime Val; raw MAE $6.4125$ post-NaN filter)}\\[2pt]
\begin{tabular*}{\textwidth}{@{\extracolsep{\fill}}llccccc@{}}
\toprule
Zero-Shot VLM & Visual Prompt & Parse OK & Val Raw MAE & Corr.\ MAE & $\Delta$MAE & Harm (\%) \\
\midrule
MedGemma-1.5-4B~\cite{sellergren2026medgemma15}$^{\dagger}$ & \texttt{proto\_diff\_pair} & 49.2\% & 6.8382$^{\dagger}$ & 7.1346 & $-0.2965$ & 40.8 \\
Qwen3-VL-8B~\cite{bai2025qwen3vl} & \texttt{proto\_diff\_pair} & 100.0\% & 6.4125 & 6.8066 & $-0.3941$ & 54.2 \\
LLaVA-Video-7B~\cite{zhang2024llavavideo} & 8-frame video overlay & 100.0\% & 6.4125 & 6.4063 & $+0.0063$ & 35.8 \\
Qwen3-VL-32B-FP8 & \texttt{proto\_diff\_pair} & 100.0\% & 6.4125 & 7.5046 & $-1.0921$ & 51.7 \\
\bottomrule
\end{tabular*}
\begin{tablenotes}[flushleft]
\scriptsize
\item Panel~(a): MC3 = 3-class severity; gNet = net risk-gated correct flips; AURC = area under risk--coverage curve. Track~1A AURC: Scalar-MLP $3.788$, 3D-Shuf $3.787$, 3D-Attr Real $3.617$. Paired Track~3 blind-test LoRA-VLM AURC (Table~\ref{tab:error_predictor}): MedGemma $4.9650$, Qwen3-VL-8B $5.3895$, LLaVA-Video-7B $4.6306$. Panel~(c): $^{\dagger}$Earliest MedGemma run preceded removal of 18 corrupt crops ($6.8382$ vs.\ $6.4125$).
\end{tablenotes}
\end{threeparttable}
\end{table*}

\subsection{Controlled Mechanistic Addenda}
\label{app:mech_addenda}
Table~\ref{tab:app_overnight_extra} collects four controlled experiments cited in Section~\ref{subsec:exp_core_pillars}: memory sub-layer isolation, a same-backbone LLM-as-verifier replacement for the harness, frame-budget and prompt sensitivity of direct VLMs, and a three-seed controlled ablation of wrong-rollout advantage clamping.

\begin{table*}[!t]
\centering
\caption{Controlled mechanistic addenda (zero missing predictions): (a)~Memory sub-layer isolation on Qwen3-8B (no Harness / no Attr-Corr), (b)~same-backbone Qwen3-8B LLM-as-Verifier vs.\ deterministic Harness, (c)~16-frame and full capability-coverage prompt sensitivity for Direct VLMs, and (d)~strictly controlled 3-seed $\text{PAW}=0$ RL ablation (seeds $42,43,44$; fixed \texttt{upd20} + Attr-Corr).}
\label{tab:app_overnight_extra}
\scriptsize
\renewcommand{\arraystretch}{0.92}
\setlength{\tabcolsep}{2.2pt}
\begin{threeparttable}
\noindent\parbox{\textwidth}{\textbf{(a) Memory sub-layer isolation on Qwen3-8B Study Test ($N{=}242$) and Study Val ($N{=}210$)}}\\[2pt]
\begin{tabular*}{\textwidth}{@{\extracolsep{\fill}}lcccccc@{}}
\toprule
Arm (No Harness, No Attr-Corr) & Test Strict (\%) & 95\% CI & Test Ans.\ (210) & Test $U_*$ (32) & Val Strict (\%) & Test Calls / Inv \\
\midrule
Qwen3-8B Base & 56.61 & [49.53, 63.59] & 54.76 & 68.8\% (22/32) & 53.33 & 3.10 / 0.56 \\
+ Memory (Calibration Priors only) & 59.50 & [53.17, 65.27] & 60.48 & 53.1\% (17/32) & 55.71 & 2.90 / 0.51 \\
+ Memory (Workflow Playbooks only) & 64.05 & [57.33, 70.87] & 60.00 & 90.6\% (29/32) & \textbf{63.33} & 2.27 / 0.16 \\
\rowcolor{oursrow}\textbf{+ Memory (Full: Priors + Playbooks)} & \textbf{67.36} & [60.69, 73.33] & \textbf{63.81} & \textbf{90.6\%} (29/32) & 62.86 & \textbf{2.18 / 0.16} \\
\bottomrule
\end{tabular*}

\vspace{3pt}
\noindent\parbox{\textwidth}{\textbf{(b) Same-backbone Qwen3-8B LLM-as-Verifier vs.\ Deterministic Verification Harness on Study Test ($N{=}242$)}}\\[2pt]
\begin{tabular*}{\textwidth}{@{\extracolsep{\fill}}lcccccc@{}}
\toprule
Verifier Configuration & Test Strict (\%) & 95\% CI & Test Ans.\ (210) & Test $U_*$ (32) & Calls / Inv & Mean Wall Time \\
\midrule
Qwen3-8B + Memory (no verifier) & 67.36 & [60.69, 73.33] & 63.81 & 90.6\% (29/32) & 2.18 / 0.16 & 1.90\,s \\
Qwen3-8B + Memory + LLM-as-Verifier & 66.94 & [59.82, 73.45] & 63.33 & 90.6\% (29/32) & 2.24 / 0.18 & 2.76\,s ($+41\%$) \\
\rowcolor{bestrow}\textbf{Qwen3-8B + Memory + Deterministic Harness} & \textbf{68.60} & [62.25, 74.64] & \textbf{63.81} & \textbf{100.0\%} (32/32) & \textbf{2.26 / 0.14} & \textbf{1.96\,s} \\
\bottomrule
\end{tabular*}

\vspace{3pt}
\noindent\parbox{\textwidth}{\textbf{(c) Direct VLM 16-frame budget and full capability-coverage prompt sensitivity on Study ($N{=}242$) and Single-Video ($N{=}140$)}}\\[2pt]
\begin{tabular*}{\textwidth}{@{\extracolsep{\fill}}lcccccc@{}}
\toprule
VLM \& Prompt / Frame Budget & Study Ans / Strict & Study $U_*$ (32) & Study FalseAbs & SV Ans / Strict & SV $U_*$ (20) & SV FalseAbs \\
\midrule
MedGemma-1.5-4B (8f, feasibility ref.) & 31.40 / 0.00 & 0.0\% (0/32) & 0.5\% (1) & 35.71 / 0.00 & 0.0\% (0/20) & 0.0\% (0) \\
MedGemma-1.5-4B (16f, feasibility) & 33.06 / 0.00 & 0.0\% (0/32) & 0.0\% (0) & 37.86 / 0.71 & 5.0\% (1/20) & 0.0\% (0) \\
MedGemma-1.5-4B (16f + full coverage prompt) & 30.17 / 1.24 & 9.4\% (3/32) & 1.9\% (4) & 37.14 / 0.71 & 5.0\% (1/20) & 5.8\% (7) \\
Qwen3-VL-8B (8f, feasibility ref.) & 17.77 / 3.72 & 28.1\% (9/32) & 49.0\% (103) & 21.43 / 7.14 & 50.0\% (10/20) & 65.0\% (78) \\
Qwen3-VL-8B (16f, feasibility) & 20.25 / 2.89 & 21.9\% (7/32) & 43.8\% (92) & 23.57 / 4.29 & 30.0\% (6/20) & 55.8\% (67) \\
Qwen3-VL-8B (16f + full coverage prompt) & 20.66 / 13.22 & 100.0\% (32/32) & 73.8\% (155) & 20.71 / 14.29 & 100.0\% (20/20) & 82.5\% (99) \\
\bottomrule
\end{tabular*}

\vspace{3pt}
\noindent\parbox{\textwidth}{\textbf{(d) Strictly controlled 3-seed $\text{PAW}=0$ RL ablation (same OOF-SFT init., curriculum, replay; seeds $42,43,44$; mean $\pm$ SD)}}\\[2pt]
\begin{tabular*}{\textwidth}{@{\extracolsep{\fill}}lcccccc@{}}
\toprule
Controlled RL Condition ($3/3$ seeds) & Train $\text{PAW}>0$ & Study Val Strict & Study Test Strict & Test Ans.\ (210) & ConcG (Test) & Test Inv \\
\midrule
Unguarded ($\text{PAW}>0$ allowed, no cross-check) & $18.0$ ($0/23/31$) & $72.54 \pm 0.55\%$ & $70.39 \pm 0.24\%$ & $65.87 \pm 0.28\%$ & $98.62 \pm 0.24\%$ & $0.023 \pm 0.006$ \\
$\text{PAW}=0$ guard only (no cross-check) & \textbf{$0.0$} ($0/0/0$) & $71.75 \pm 0.99\%$ & $70.25 \pm 0.00\%$ & $65.71 \pm 0.00\%$ & $98.49 \pm 0.24\%$ & $0.027 \pm 0.006$ \\
\rowcolor{bestrow}\textbf{$\text{PAW}=0$ + group-cond.\ cross-check} & \textbf{$0.0$} ($0/0/0$) & \textbf{$73.02 \pm 0.73\%$} & \textbf{$70.39 \pm 0.24\%$} & \textbf{$65.87 \pm 0.28\%$} & $98.49 \pm 0.24\%$ & $0.030 \pm 0.000$ \\
\bottomrule
\end{tabular*}
\begin{tablenotes}[flushleft]
\scriptsize
\item Panel~(a) nominal exact McNemar tests on Study Test: Base$\rightarrow$Playbooks $+7.44$\,pp ($20{:}2$, $p{=}0.00012$); Priors$\rightarrow$Full $+7.85$\,pp ($23{:}4$, $p{=}0.00031$); Playbooks$\rightarrow$Full $+3.31$\,pp ($18{:}10$, $p{=}0.185$). Panel~(b): Deterministic Harness vs.\ LLM-Verifier $+1.65$\,pp ($12{:}8$, $p{=}0.503$), with $100\%$ vs.\ $90.6\%$ refusal and $1.96$\,s vs.\ $2.76$\,s mean latency. Panel~(d) per-seed Study Val / Test Strict for $\text{PAW}=0$ + cross-check: s42 $73.81/70.25\%$, s43 $72.86/70.66\%$, s44 $72.38/70.25\%$.
\end{tablenotes}
\end{threeparttable}
\end{table*}

% ------------------------------------------------------------------------------
\section{Safety, Unseen Generalization, and End-to-End Runtime}
\label{app:safety_runtime}
\begin{figure*}[!t]
\centering
\includegraphics[width=\textwidth]{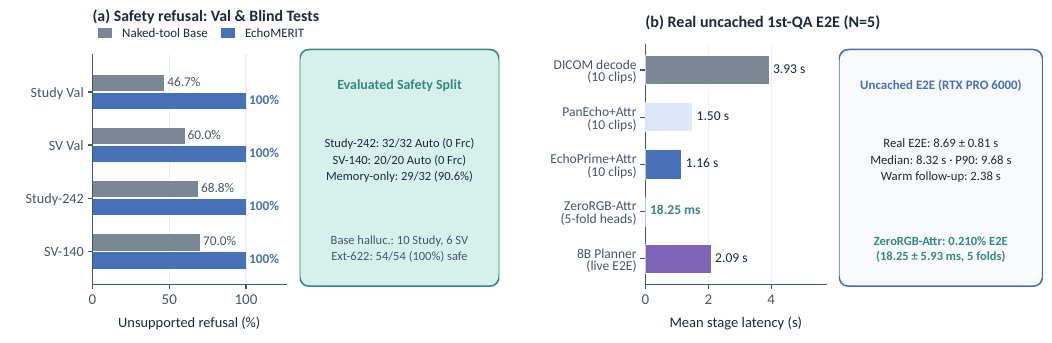}
\caption{Safety abstention decomposition and H4 real uncached end-to-end runtime. (a)~Unsupported refusal across Val and Blind Tests: naked-tool Base answers $10/32$ (Study-242) and $6/20$ (SV-140) unsupported items and Memory-only reaches $90.6\%$ ($29/32$), whereas full \sysname{} reaches $100\%$ autonomous refusal ($0$ forced truncations; $54/54$ external refusal). (b)~Real uncached end-to-end first-question wall time on $5$ unseen $10$-clip DICOM studies (NVIDIA RTX PRO~6000 Blackwell Max-Q, \texttt{precache=False}): $8.69\pm 0.81$\,s total ($18.25\pm 5.93$\,ms or $0.210\%$ for 5-fold ZeroRGB-Attr, $2.09\pm 0.40$\,s live Qwen3-8B planning).}
\label{fig:app_safety_latency}
\end{figure*}

Table~\ref{tab:app_safety_latency} and Fig.~\ref{fig:app_safety_latency} decompose the safety, generalization, and runtime results summarized in Section~\ref{subsec:exp_core_pillars}. Panel~(a) separates autonomous refusals from Gate~1/1b forced truncations and reports false abstention on answerable items. Every full \sysname{} configuration refuses all unsupported test items autonomously, so the forced-abstention backup is never needed on these subsets. Panel~(b) decomposes the held-out unseen tasks; because these subsets are small, they should be read descriptively. Panel~(c) audits the same-weight answer flips induced by tool correction, and panel~(d) profiles real uncached end-to-end runtime on unseen DICOM studies, where decoding and encoder passes dominate and the five-fold \emph{ZeroRGB-Attr} forward pass accounts for about $0.2\%$ of wall time.

% Appendix Table A-VIII (tab:app_safety_latency): Safety, Held-Out Split, Flip Audit & Composed H4 Runtime
\begin{table*}[!t]
\centering
\caption{Clinical safety decomposition, held-out unseen task split, tool-correction flip audit (including both frontier families on Study Val), and H4 real uncached end-to-end first-question runtime on an NVIDIA RTX PRO~6000 Workstation GPU.}
\label{tab:app_safety_latency}
\scriptsize
\renewcommand{\arraystretch}{0.92}
\setlength{\tabcolsep}{2.1pt}
\begin{threeparttable}

\textbf{(a) Safety and abstention audit (H1; unsupported refusal $U_*$ autonomous vs.\ Gate~1/1b forced truncation, and false abstention)}\\[2pt]
\begin{tabular*}{\textwidth}{@{\extracolsep{\fill}}lccccccc@{}}
\toprule
Arm & \makecell{Val $U_*$\\St/SV (\%)} & \makecell{Study-242 $U_*$\\Auto/Frc (32)} & \makecell{Study-242\\FalseAbs} & \makecell{SV-140 $U_*$\\Auto/Frc (20)} & \makecell{SV-140\\FalseAbs} & \makecell{Test $U_*$\\Halluc.} & \makecell{Ext-622\\Uncov.\ (54)} \\
\midrule
MedGemma-1.5-4B (VLM) & --- & 0 / 0 (0.0\%) & 0.48\% (1) & 0 / 0 (0.0\%) & 0.00\% (0) & 32 / 20 & 0.0\% (0) \\
LLaVA-Video-7B (VLM) & --- & 0 / 0 (0.0\%) & 0.48\% (1) & 0 / 0 (0.0\%) & 1.67\% (2) & 32 / 20 & 0.0\% (0) \\
Qwen3-VL-8B (VLM)$^\star$ & --- & 9 / 0 (28.1\%) & 49.05\% (103) & 10 / 0 (50.0\%) & 65.00\% (78) & 23 / 10 & --- \\
Qwen3-VL-32B-FP8 (VLM)$^\star$ & --- & 1 / 0 (3.1\%) & 41.90\% (88) & 3 / 0 (15.0\%) & 53.33\% (64) & 31 / 17 & --- \\
\midrule
Qwen3-8B Base & 46.7 / 60.0 & 22 / 0 (68.8\%) & 0.00\% (0) & 14 / 0 (70.0\%) & 0.83\% (1) & 10 / 6 & \textbf{100.0\%} (54) \\
Qwen3-8B + Memory only & 86.7 / --- & 29 / 0 (90.6\%) & 0.00\% (0) & --- & --- & 3 / --- & --- \\
Zero-Shot (H+M / +Attr-Corr) & \textbf{100 / 100} & \textbf{32 / 0 (100\%)} & \textbf{0.00\% (0)} & \textbf{20 / 0 (100\%)} & \textbf{0.00\% (0)} & \textbf{0 / 0} & \textbf{100.0\%} (54) \\
OOF-SFT (+Attr-Corr) & \textbf{100 / 100} & \textbf{32 / 0 (100\%)} & \textbf{0.00\% (0)} & \textbf{20 / 0 (100\%)} & \textbf{0.00\% (0)} & \textbf{0 / 0} & \textbf{100.0\%} (54) \\
EchoMERIT-RL (+Attr-Corr) & \textbf{100 / 100} & \textbf{32 / 0 (100\%)} & \textbf{0.00\% (0)} & \textbf{20 / 0 (100\%)} & 0.83\% (1) & \textbf{0 / 0} & \textbf{100.0\%} (54) \\
DeepSeek / Qwen3.8 (All) & \textbf{100 / 100} & \textbf{32 / 0 (100\%)} & \textbf{0.00\% (0)} & \textbf{20 / 0 (100\%)} & \textbf{0.00\% (0)} & \textbf{0 / 0} & \textbf{100.0\%} (54) \\
\bottomrule
\end{tabular*}

\vspace{2pt}
\textbf{(b) Held-out unseen task decomposition (H2; $6{+}2$ tasks excluded from planner SFT/GRPO trajectories; small subsets descriptive)}\\[2pt]
\begin{tabular*}{\textwidth}{@{\extracolsep{\fill}}lccccccc@{}}
\toprule
Arm & \makecell{Val Seen\\(183)} & \makecell{Val Uns-\\Ans (18)} & \makecell{Val Uns-\\Abs (9)} & \makecell{Study-242\\Uns-Ans (21)} & \makecell{Study-242\\Uns-Abs (7)} & \makecell{Text-500\\Uns (100)} & \makecell{Study-242\\Calls / Inv} \\
\midrule
Qwen3-8B Base & --- & 55.6 & --- & 47.6 & 71.4 & 25.0 & 3.10 / 0.56 \\
Zero-Shot (H+M / +Attr-Corr) & 62.3 / 65.0 & 72.2 & \textbf{100.0} & 57.1 / 61.9 & \textbf{100.0} & 66.0 & 2.26/0.14 $\rightarrow$ 2.12/0.13 \\
OOF-SFT (Scalar / Attr-Corr) & 71.6 & 72.2 & \textbf{100.0} & \textbf{71.4} / 61.9 & \textbf{100.0} & \textbf{67.0} & 1.82/0.02 \;$|$\; 1.80/0.02 \\
EchoMERIT-RL (Scalar / Attr-Corr) & 72.1 & 72.2 & \textbf{100.0} & \textbf{71.4} / 61.9 & \textbf{100.0} & \textbf{67.0} & \textbf{1.78 / 0.01} \\
DeepSeek (Base$\rightarrow$+H+M$\rightarrow$+Attr) & --- & 66.7 & \textbf{100.0} & 52.4 / 57.1 / 66.7 & \textbf{100.0} & --- & 3.07/0.13 $\rightarrow$ 2.42/0.09 \\
Qwen3.8 (Base$\rightarrow$+H+M$\rightarrow$+Attr) & --- & --- & --- & 52.4 / 57.1 / 66.7 & \textbf{100.0} & \textbf{100.0} & 3.10/0.16 $\rightarrow$ 2.69/0.11 \\
\bottomrule
\end{tabular*}

\vspace{2pt}
\textbf{(c) Tool-correction flip audit (H3; same-weight Uncorrected$\rightarrow$Corrector; nominal QA-level exact McNemar $p$)}\\[2pt]
\begin{tabular*}{\textwidth}{@{\extracolsep{\fill}}lccccc@{}}
\toprule
Cohort \& Policy (Backend) & W$\rightarrow$R & R$\rightarrow$W & Net & Exact McNemar $p$ & $U_*$ Harm \\
\midrule
Study Val 210: DeepSeek +H+M$\rightarrow$+Attr-Corr & 25 & 8 & $+17$ & $0.00455$ & 0.0\% \\
Study Val 210: Qwen3.8 +H+M$\rightarrow$+Attr-Corr & 22 & 7 & $+15$ & $0.00813$ & 0.0\% \\
Study Val 210: OOF-SFT / Harness-RL / RL (None$\rightarrow$Attr) & 18 / 24 / 20 & 8 / 11 / 8 & $+10 / +13 / +12$ & $0.0755$ / $0.0410$ / $0.0357$ & 0.0\% \\
Study-242: OOF-SFT / RL (None$\rightarrow$Scalar) & 17 / 18 & 15 / 15 & $+2 / +3$ & $0.8601$ / $0.7283$ (ns) & 0.0\% \\
Study-242: OOF-SFT / RL (None$\rightarrow$Attr-Corr) & 17 / 18 & 17 / 17 & $0 / +1$ & $1.0000$ / $1.0000$ (ns) & 0.0\% \\
SV-140: ZS / OOF-SFT / RL (None$\rightarrow$Attr-Corr) & 14 / 20 / 18 & 7 / 11 / 9 & $+7 / +9 / +9$ & $0.189$ / $0.150$ / $0.122$ (trend) & 0.0\% \\
SV-140: EchoMERIT-RL (None$\rightarrow$Scalar) & 20 & 9 & $+11$ & $0.0614$ (CI $[+1.28,+14.78]$) & 0.0\% \\
SV-140: Qwen3.8 +H+M$\rightarrow$+Attr-Corr & 13 & 8 & $+5$ & $0.3833$ (ns; vs.\ Base $0.0060$) & 0.0\% \\
\bottomrule
\end{tabular*}

\vspace{2pt}
\textbf{(d) Real uncached end-to-end 1st-QA runtime on unseen $10$-clip DICOM studies (H4; NVIDIA RTX PRO~6000 Blackwell Max-Q, $N{=}5$, \texttt{precache=False})}\\[2pt]
\begin{tabular*}{\textwidth}{@{\extracolsep{\fill}}lcccc@{}}
\toprule
Pipeline Stage & Mean $\pm$ SD & Median & P90 & Share of E2E Total \\
\midrule
1. Real DICOM multi-frame decode (10 clips/study, uncached) & $3.93 \pm 0.78$\,s & $3.82$\,s & $4.81$\,s & $45.2\%$ \\
2. PanEcho (40 heads) + attribution heatmap extraction (10 clips) & $1.50 \pm 0.35$\,s & $1.33$\,s & $1.86$\,s & $17.3\%$ \\
3. EchoPrime (21 phenotypes) + heatmaps + retrieval (10 clips) & $1.16 \pm 0.28$\,s & $1.02$\,s & $1.44$\,s & $13.3\%$ \\
\rowcolor{oursrow}\textbf{4. 5-Fold ZeroRGB-Attr forward (all queried heads, mean $1.2$)} & \textbf{$18.25 \pm 5.93$\,ms} & \textbf{$15.35$\,ms} & \textbf{$24.22$\,ms} & \textbf{$0.210\%$} \\
5. Qwen3-8B (EchoMERIT-RL) live same-GPU multi-turn planning & $2.09 \pm 0.40$\,s & $2.06$\,s & $2.50$\,s & $24.1\%$ \\
\midrule
\rowcolor{bestrow}\textbf{Real uncached end-to-end 1st-QA wall time ($N{=}5$ unseen studies)} & \textbf{$8.69 \pm 0.81$\,s} & \textbf{$8.32$\,s} & \textbf{$9.68$\,s} & \textbf{$100.0\%$} \\
\textit{Cross-check: $N{=}10$ preloaded visual ($6.46$\,s) + $N{=}242$ cached plan ($2.38$\,s)} & $8.85$\,s (warm $2.38$\,s) & $1.91$\,s (warm) & $4.27$\,s (warm) & --- \\
\bottomrule
\end{tabular*}

\begin{tablenotes}[flushleft]
\scriptsize
\item Panel~(a): $^\star$Qwen3-VL-8B/32B abstain on $435/277$ of 622 external queries overall. Exact two-sided $95\%$ Clopper--Pearson CIs for $100\%$ refusal subsets: $32/32$ $[89.1,100.0]\%$, $20/20$ $[83.2,100.0]\%$, $54/54$ $[93.4,100.0]\%$. Memory-only Study Val $U_*$ is $26/30$ ($86.7\%$); Test $U_*$ is $29/32$ ($90.6\%$). SV-140 and Ext-622 frontier entries report Qwen3.8 (DeepSeek Ext-622 full+Attr also attains $54/54$ [$100\%$] refusal; Table~\ref{tab:app_mimicechoqa}). Panel~(d): Measured end-to-end in a single process on NVIDIA RTX PRO~6000 Blackwell Max-Q Workstation Edition (98\,GB, CUDA AMP, batch $1$, \texttt{precache\_used=False}), closely corroborating the earlier $N{=}10$ composed estimate ($8.85$\,s; $15.53$\,ms for 1 pre-cached EF item).
\end{tablenotes}
\end{threeparttable}
\end{table*}

% ------------------------------------------------------------------------------
\section{External MIMICEchoQA Audit and Diagnostic Case Studies}
\label{app:external_cases}
This appendix expands the external benchmark note of Section~\ref{subsec:exp_error_predictor} and illustrates runtime mechanics with four diagnostic episodes.

\subsection{External MIMICEchoQA Audit}
\label{app:external_audit}
To evaluate zero-shot transfer and benchmark validity without patient contamination, all $582$ MIMICEchoQA patients~\cite{thapa2025openbiomedvid} were locked exclusively to Test ($N{=}622$). An objective audit of MIMICEchoQA uncovers four structural shortcut and feasibility issues: (1)~a blind majority-position \emph{Always-B} baseline achieves $57.40\%$ ($357/622$), exceeding MedGemma-1.5-4B ($54.02\%$), LLaVA-Video-7B ($41.80\%$), Qwen3-VL-32B ($25.24\%$), and published $120\text{B}$ Echo-CoPilot~v4 ($54.0\pm 0.7\%$~\cite{heidari2025echocopilot}); (2)~across the $320$ four-choice items, answer letters skew sharply ($A$: $2.5\%$ [$8/320$], $B$: $62.8\%$ [$201/320$], $C$: $28.1\%$ [$90/320$], $D$: $6.6\%$ [$21/320$]; $B{+}C = 90.9\%$), including $63/63$ LVEF items ($0$ in A/D) and $81/81$ MR/AR/LVH severity items where \emph{Normal} never appears (because severity templates trigger only when pathology is reported), while yes/no queries heavily favor \emph{No} (e.g., $29/29$ AS and $18/18$ MVP); (3)~$54$ queries ($8.7\%$) target uncovered structures (ASD/PFO, MVP, pulmonic valve); and (4)~$21+$ Left Atrium queries ask for LAVI despite MIMICEchoQA omitting patient height/weight/BSA metadata required for physical calculation. Table~\ref{tab:app_mimicechoqa} reports full stratified metrics across all direct VLMs, local $8\text{B}$, Qwen3.8, and DeepSeek arms together with exact anatomical category counts. Under our auditable protocol, every \sysname{} backbone maintains $100\%$ refusal on the $54$ uncovered items, and plug-and-play modules lift Qwen3.8 by $6.43$ points ($p{=}2.0\times 10^{-6}$). In contrast, the trained $8\text{B}$ planner trails its base agent in overall ($37.62\%$ vs.\ $49.04\%$) and answered-item accuracy despite about nine times fewer invalid calls, and 2D \emph{Attr-Corr} lowers accuracy for both backbones. Forced-choice models can thus achieve high raw accuracy from answer priors alone, and we treat prior and template shift as an open adaptation problem.

\begin{table*}[!t]
\centering
\caption{External zero-shot MIMICEchoQA~\cite{thapa2025openbiomedvid} audit ($622$ QA, $582$ patients; dual-tool PE+EP + shared-$w_0$ Attr-Corr; zero missing predictions): (a)~full arm comparison including the blind Always-B shortcut baseline, four direct VLMs, local $8\text{B}$, Qwen3.8, and DeepSeek, and (b)~exact anatomical category Strict accuracy decomposition (including uncovered items).}
\label{tab:app_mimicechoqa}
\scriptsize
\setlength{\tabcolsep}{2.4pt}
\begin{threeparttable}
\noindent\parbox{\textwidth}{\textbf{(a) Full arm comparison, blind shortcut baseline, covered vs.\ uncovered refusal, and Pericardium conservatism audit}}\\[2pt]
\begin{tabular*}{\textwidth}{@{\extracolsep{\fill}}lccccccc@{}}
\toprule
Arm & \makecell{All 622\\Ans / Strict} & \makecell{95\% CI\\(Strict)} & \makecell{Covered\\($N=568$)} & \makecell{Ans-Only\\($n$ ans.)} & \makecell{Uncovered 54\\Refusal (\%)} & \makecell{Pericardium 82\\Acc \% (Abs.)} & \makecell{Calls / Inv} \\
\midrule
\rowcolor{sectbg}Blind Always-B Shortcut (no video/reasoning) & \textbf{57.40} / 0.00 & [0.00, 0.00] & --- & 57.40 (622) & 0.0 & --- & --- \\
MedGemma-1.5-4B (Direct VLM) & 54.02 / 0.00 & [0.00, 0.00] & --- & 54.02 (622) & 0.0 & --- & --- \\
LLaVA-Video-7B (Direct VLM) & 41.80 / 0.00 & [0.00, 0.00] & --- & 41.80 (622) & 0.0 & --- & --- \\
Qwen3-VL-8B (Direct VLM)$^\star$ & 13.18 / 0.00 & [0.00, 0.00] & --- & --- (187) & --- & --- & --- \\
Qwen3-VL-32B-FP8 (Direct VLM)$^\star$ & 25.24 / 0.00 & [0.00, 0.00] & --- & --- (345) & --- & --- & --- \\
Echo-CoPilot~v4 (120B; published ref.-only)~\cite{heidari2025echocopilot} & 54.0$\pm$0.7 / --- & --- & --- & --- & --- & --- & --- \\
\midrule
Qwen3-8B No-Tools (no image) & 21.54 / 0.00 & [0.00, 0.00] & 18.66 & 24.95 (537) & 42.6 & 39.0 (4) & 0.00 / 0.00 \\
Qwen3-8B Base (naked tools) & \textbf{49.04 / 49.04} & [45.06, 53.03] & \textbf{53.70} & \textbf{53.98} (565) & \textbf{100.0} & 67.1 (0) & 3.15 / 0.83 \\
Zero-Shot (H+M) & 47.11 / 47.11 & [43.22, 50.90] & 51.58 & 51.95 (564) & \textbf{100.0} & 67.1 (0) & 3.18 / 0.61 \\
Zero-Shot + Attr-Corr & 42.44 / 42.44 & [38.61, 46.25] & 46.48 & 46.98 (562) & \textbf{100.0} & 65.9 (0) & 2.06 / 0.41 \\
OOF-SFT (Uncorr.) & 33.92 / 33.92 & [30.21, 37.74] & 37.15 & 40.89 (516) & \textbf{100.0} & 20.7 (46) & 1.72 / 0.13 \\
OOF-SFT + Attr-Corr & 37.62 / 37.62 & [33.92, 41.38] & 41.20 & 42.16 (555) & \textbf{100.0} & 67.1 (1) & 1.42 / 0.10 \\
EchoMERIT-RL (Uncorr.) & 39.07 / 39.07 & [35.38, 42.93] & 42.78 & 43.71 (556) & \textbf{100.0} & 59.8 (8) & 1.62 / 0.13 \\
\rowcolor{oursrow}EchoMERIT-RL + Attr-Corr (8B Final) & 37.62 / 37.62 & [33.82, 41.42] & 41.20 & 41.79 (560) & \textbf{100.0} & 65.9 (0) & \textbf{1.42 / 0.09} \\
\midrule
Qwen3.8 Base (naked tools) & 39.87 / 39.71 & [35.85, 43.47] & 43.66 & 45.01 (551) & \textbf{100.0} & 64.6 (1) & 3.47 / \textbf{0.06} \\
\rowcolor{bestrow}\textbf{Qwen3.8 + H+M (no corr.)} & 46.14 / 46.14 & [42.22, 49.84] & 50.53 & 51.81 (554) & \textbf{100.0} & \textbf{68.3} (0) & 3.71 / 0.13 \\
Qwen3.8 + Attr-Corr & 42.28 / 42.28 & [38.47, 46.03] & 46.30 & 47.39 (555) & \textbf{100.0} & 67.1 (1) & 3.52 / 0.12 \\
DeepSeek + Full + Attr-Corr & 37.94 / 37.94 & [34.25, 41.73] & 41.55 & --- & \textbf{100.0} & 65.8 (2) & 3.90 / 0.15 \\
\bottomrule
\end{tabular*}

\vspace{3pt}
\noindent\parbox{\textwidth}{\textbf{(b) Exact anatomical category breakdown (overall Strict correct count and \%, denominator $N$ includes uncovered items refused by all tool arms)}}\\[2pt]
\begin{tabular*}{\textwidth}{@{\extracolsep{\fill}}lcccccc@{}}
\toprule
Anatomical Category & $N$ & Uncov.\ $n$ & 8B Base & 8B Final (RL+Attr) & Qwen3.8 Base & Qwen3.8 + H+M \\
\midrule
Left Ventricle & 162 & 1 & 76 (46.9\%) & 58 (35.8\%) & 61 (37.6\%) & 76 (46.9\%) \\
Aortic Valve & 132 & 1 & 84 (63.6\%) & 64 (48.5\%) & 71 (53.8\%) & 80 (60.6\%) \\
Mitral Valve & 87 & 22 & 40 (46.0\%) & 18 (20.7\%) & 20 (23.0\%) & 23 (26.4\%) \\
Pericardium & 82 & 0 & 55 (67.1\%) & 54 (65.8\%) & 53 (64.6\%) & 56 (68.3\%) \\
Left Atrium & 38 & 1 & 17 (44.7\%) & 6 (15.8\%) & 14 (36.8\%) & 17 (44.7\%) \\
Atrial Septum (uncovered) & 24 & 24 & 0 (0.0\%) & 0 (0.0\%) & 0 (0.0\%) & 0 (0.0\%) \\
Tricuspid Valve & 23 & 0 & 12 (52.2\%) & 7 (30.4\%) & 9 (39.1\%) & 14 (60.9\%) \\
Right Ventricle & 23 & 0 & 8 (34.8\%) & 9 (39.1\%) & 8 (34.8\%) & 9 (39.1\%) \\
Other 6 structures$^\dagger$ & 51 & 5 & 13 (25.5\%) & 18 (35.3\%) & 11 (21.6\%) & 12 (23.5\%) \\
\midrule
\textbf{Total (All MIMICEchoQA)} & \textbf{622} & \textbf{54} & \textbf{305 (49.04\%)} & \textbf{234 (37.62\%)} & \textbf{247 (39.71\%)} & \textbf{287 (46.14\%)} \\
\bottomrule
\end{tabular*}
\begin{tablenotes}[flushleft]
\scriptsize
\item Uncovered $54$ items ($8.7\%$): Atrial Septum / ASD--PFO ($24$), Mitral Valve / MVP ($22$), Pulmonic Valve ($4$), Aorta/AV/LA/LV ($4$). $^\dagger$Other 6 structures: Aorta ($18$, uncov.\ $1$), Pulmonary Artery ($12$), Right Atrium ($11$), Wall Motion ($5$), Pulmonic Valve ($4$, uncov.\ $4$), IVC ($1$). $^\star$Qwen3-VL-8B/32B abstain on $435/277$ of $622$ queries overall. Nominal paired exact McNemar tests: 8B Base$\rightarrow$ZS H+M $-1.93$\,pp ($p{=}0.219$, ns); ZS H+M$\rightarrow$+Attr $-4.66$\,pp ($25{:}54$, $p{=}0.0015$); OOF-SFT Uncorr.$\rightarrow$+Attr $+3.70$\,pp ($67{:}44$, $p{=}0.0363$, Pericardium abstentions $46\!\rightarrow\!1$); OOF-SFT Uncorr.$\rightarrow$RL Uncorr.\ $+5.14$\,pp ($37{:}5$, $p{<}10^{-5}$, $46\!\rightarrow\!8$); RL Uncorr.$\rightarrow$+Attr $-1.45$\,pp ($33{:}42$, $p{=}0.3557$, ns); Qwen3.8 Base$\rightarrow$+H+M $+6.43$\,pp ($55{:}15$, $p{=}2.0\times 10^{-6}$, CI $[+3.83,+9.08]$); Qwen3.8 +H+M$\rightarrow$+Attr $-3.86$\,pp ($25{:}49$, $p{=}0.0071$).
\end{tablenotes}
\end{threeparttable}
\end{table*}

\begin{figure*}[!t]
\centering
\includegraphics[width=\textwidth]{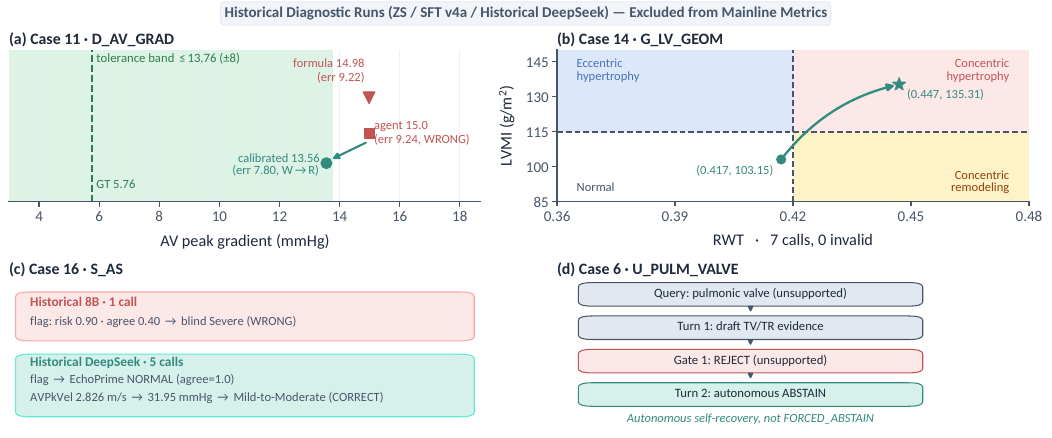}
\caption{Qualitative anatomy of four diagnostic episodes (excluded from mainline metrics). (a)~Case~11 (\texttt{D\_AV\_GRAD}): calibrated \texttt{AVPkVel} moves Bernoulli gradient from $15.0$ to $13.56$\,mmHg inside the $\pm 8$\,mmHg band. (b)~Case~14 (\texttt{G\_LV\_GEOM}): $7$-call cubic cascade crossing $\text{RWT}=0.42$ and $\text{LVMI}=115$\,g/m$^2$. (c)~Case~16 (\texttt{S\_AS}): split-consensus cross-checking recovering Mild-to-Moderate. (d)~Case~6 (\texttt{U\_PULM\_VALVE}): Gate~1 rejection followed by autonomous abstention.}
\label{fig:app_cases}
\end{figure*}

\subsection{Qualitative Diagnostic Case Studies}
\label{app:cases}
Four episodes from historical diagnostic runs, excluded from all mainline metrics, concretely illustrate runtime mechanics (Fig.~\ref{fig:app_cases}).
\begin{itemize}
\item \textbf{Case~6:} \texttt{U\_PULM\_VALVE}, Gate~1 rejection followed by self-recovery. On a pulmonic-valve query, the planner initially substituted tricuspid evidence; Gate~1 rejected the draft, and the planner autonomously self-recovered to \texttt{ABSTAIN}.
\item \textbf{Case~11:} \texttt{D\_AV\_GRAD}, dual-track Bernoulli threshold crossing. Against reference $5.76$\,mmHg ($\pm 8.0$\,mmHg benchmark tolerance, not clinician acceptance), raw \texttt{AVPkVel} $=1.935$\,m/s yields $14.98$\,mmHg (wrong at $15.0$\,mmHg). Calibrating \texttt{AVPkVel} to $1.841$\,m/s feeds \texttt{calculate} to yield $13.56$\,mmHg (error $7.80$\,mmHg, $\text{W}\!\rightarrow\!\text{R}$).
\item \textbf{Case~14:} \texttt{G\_LV\_GEOM}, seven-step cubic cascade. Across $7$ tool calls ($0$ invalid)~\cite{lang2015chamber}, joint calibration of \texttt{LVIDd}, \texttt{IVSd}, and \texttt{LVPWd} propagates through Devereux LV mass ($210.3\!\rightarrow\!275.9$\,g), \texttt{LVMI} ($103.2\!\rightarrow\!135.3$\,g/m$^2$), and \texttt{RWT} ($0.417\!\rightarrow\!0.447$), flipping \texttt{Normal geometry} to \texttt{Concentric hypertrophy}.
\item \textbf{Case~16:} \texttt{S\_AS}, split-consensus cross-checking illustration. When ensemble disagreement surfaced (\texttt{fold\_agree=0.40}, risk $0.90$), a historical single-call $8\text{B}$ run emitted \texttt{Severe}, whereas a historical DeepSeek run cross-checked EchoPrime, \texttt{AVPkVel} ($2.826$\,m/s), and Bernoulli gradient ($31.95$\,mmHg)~\cite{baumgartner2017aortic} over five calls to recover \texttt{Mild-to-Moderate} (a cross-model illustration, not a causal comparison).
\end{itemize}

\end{document}